\documentclass[10pt,twocolumn,letterpaper]{article}

\usepackage[pagenumbers]{iccv} 

\usepackage{amsmath,amsfonts,bm}

\def\eqref#1{equation~\ref{#1}}

\def\1{\bm{1}}

\DeclareMathAlphabet{\mathsfit}{\encodingdefault}{\sfdefault}{m}{sl}
\SetMathAlphabet{\mathsfit}{bold}{\encodingdefault}{\sfdefault}{bx}{n}

\usepackage{amsmath}
\usepackage{amssymb}
\usepackage{dsfont}
\usepackage{xcolor}         
\usepackage{colortbl}

\newif\ifdraft
\draftfalse

\ifdraft
  \newcommand{\PF}[1]{{\color{red}{\bf PF: #1}}}
  
  \newcommand{\AF}[1]{{\color{blue}{\bf AF: #1}}}
  
  \newcommand{\CD}[1]{{\color{orange}{\bf CD: #1}}}
  
  \newcommand{\DM}[1]{{\color{green}{\bf DM: #1}}}
  
  \newcommand{\NT}[1]{{\color{violet}{\bf NT: #1}}}
  
\else
 \newcommand{\PF}[1]{}
 
 \newcommand{\AF}[1]{}
 
 \newcommand{\CD}[1]{}
 
 \newcommand{\DM}[1]{}
 
 \newcommand{\NT}[1]{}
 
\fi

\newcommand{\parag}[1]{\vspace{-2mm} \paragraph{#1}}

\newlength\mytmplen
\definecolor{iccvblue}{rgb}{0.21,0.49,0.74}
\usepackage[pagebackref,breaklinks,colorlinks,allcolors=iccvblue]{hyperref}

\def\paperID{*****} 
\def\confName{ICCV}
\def\confYear{2025}

\title{A View-consistent Sampling Method for Regularized Training of Neural Radiance Fields}

\author{Aoxiang Fan\\
Computer Vision Laboratory, EPFL\\
Switzerland\\
{\tt\small aoxiang.fan@epfl.ch}
Corentin Dumery\\
{\tt\small corentin.dumery@epfl.ch}
Nicolas Talabot\\
{\tt\small nicolas.talabot@epfl.ch}
Pascal Fua\\
{\tt\small pascal.fua@epfl.ch}
}

\author{
Aoxiang Fan$^{\scriptsize 1}$, Corentin Dumery$^{\scriptsize 1}$, Nicolas Talabot$^{\scriptsize 1}$, Pascal Fua$^{\scriptsize 1}$\\
\\
$^{1}$Computer Vision Laboratory, EPFL, Switzerland\\
{\texttt{\{aoxiang.fan, corentin.dumery, nicolas.talabot, pascal.fua\}@epfl.ch}}
}

\begin{document}

\maketitle

\begin{abstract}

    Neural Radiance Fields (NeRF) has emerged as a compelling framework for scene representation and 3D recovery. To improve its performance on real-world data, depth regularizations have proven to be the most effective ones.
    However, depth estimation models not only require expensive 3D supervision in training, but also suffer from generalization issues. As a result, the depth estimations can be erroneous in practice, especially for outdoor unbounded scenes.
    In this paper, we propose to employ view-consistent distributions instead of fixed depth value estimations to regularize NeRF training.
    Specifically, the distribution is computed by utilizing both low-level color features and high-level distilled features from foundation models at the projected 2D pixel-locations from per-ray sampled 3D points. By sampling from the view-consistency distributions, 
    an implicit regularization is imposed on the training of NeRF.
    We also utilize a depth-pushing loss that works in conjunction with the sampling technique to jointly provide effective regularizations for eliminating the failure modes. 
    Extensive experiments conducted on various scenes from public datasets demonstrate that our proposed method can generate significantly better novel view synthesis results than state-of-the-art NeRF variants as well as different depth regularization methods.

\end{abstract}    

\section{Introduction}

3D scene reconstruction from multiple images is a long-standing vision problem~\citep{Hartley00} but the recent advent of Neural Radiance Fields (NeRFs)~\citep{Mildenhall20} has delivered a significant performance boost, especially given a dense set of input images. However, in much the same way the old shape-from-shading was ill-posed~\citep{Prados05}, so is the NeRF reconstruction problem: as shown in~\citep{Zhang20f}, in the absence of explicit or implicit regularization, a set of training images can be fitted independently of the recovered geometry. This phenomenon, known as shape-radiance ambiguity, is particularly evident when the input views are not dense enough, even though using Multi-Layer Perceptrons (MLPs) for scene representation weakly regularizes the scene reconstructions~\citep{Zhang20f,Yu22}. 

Many kinds of regularizers have been proposed to improve on this, such as imposing geometric constraints~\citep{Kim22b, Niemeyer22}, training to directly predict radiance fields by using networks conditioned on image features~\citep{Chen21f, Yu21c, Wang21i}, or constraining depth~\citep{Deng22c, Wang23a, Yu22}. The first is difficult to do for complicated scenes while the second is often limited to very specific 3-view setting and not easily generalizable to unbounded scenes or scalable to more input views. The last, depth regularization, has proved to be the more widely applicable. However, it typically requires excessive expensive 3D supervision, and can produce unreliable predictions on challenging open-space scenes that produce artifacts in the final NeRF reconstruction.


\begin{figure*}[!h]
    \centering
    \includegraphics[width=0.95\linewidth]{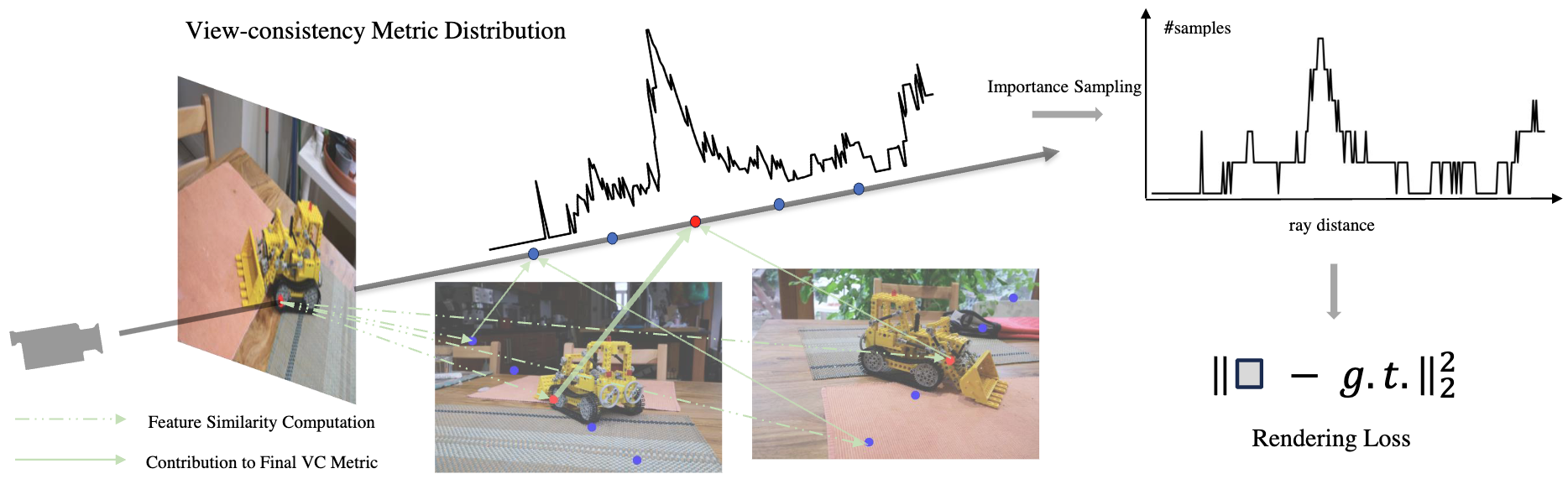}
\vspace{-3mm} %
\caption{{\bf View-consistent sampling.} Our central idea is to pre-compute a view-consistency distribution along rays and to perform importance sampling according to this distribution. As a result, the sampling will concentrate around
\textcolor{red}{surface points} instead of \textcolor{blue}{random points} in the capture volume.}
\label{fig:keyidea}
\end{figure*}

To remedy this, we propose to use view-consistency distributions per-ray instead of fixed depth predictions to implement a sampling technique along the rays that implicitly regularizes the training of NeRF, as depicted by \cref{fig:keyidea}. Specifically, we first distill geometric information from the redundant feature representations of foundation models to reduce their dimensionality and to alleviate memory requirements, while preserving the information most likely to be consistent across views and discarding the rest. Given these high-level distilled features along with low-level color features, we compute a view-consistency metric for every point along the rays and introduce an adaptive sampling scheme that favors view-consistent points, on the assumption that they are more likely to lie on a real-world surface. Furthermore, we also introduce a depth-pushing loss to force the model to favor samples that are farther away from the camera origin, which prevents the kind of background collapse artifact~\citep{Barron22} that frequently happens in NeRF reconstruction of real-world unbounded scenes. In effect, the proposed view-consistent sampling and the depth-pushing loss focus the NeRF reconstruction process on the part of the capture volume close to the true surface, thus providing implicit regularization and preventing the overfitting problem~\citep{Zhang20f}.

Our contribution is therefore a novel view-consistent sampling technique to implicitly regularize the training of NeRF. We show that our method is able to achieve significantly better novel view synthesis results compared to existing NeRF competitors with regularizations. Our implementation is based on open-source software and will be made publicly available.


\section{Related Works}
\label{sec:related}

\paragraph{NeRF Variants.}

The emergence of Neural Radiance Fields (NeRF)~\citep{Mildenhall20} is an immediate consequence of the study on Implicit Neural Representations (INR)~\citep{Tancik20,Sitzmann20,Hertz21,Mehta21},
which laid the foundation of NeRF by introducing powerful network-based scene representation models. 
NeRFs use them to effectively encode 3D scene properties and trains on posed images via a volume rendering equation that differentially relates 3D scenes to 2D images.

NeRFs  deliver outstanding image synthesis results but tend to be slow, sometimes requiring several days for a single scene. A number of accelerated approaches have therefore been proposed~\citep{Sun22,Fridovich22,Chen22f,Muller22}, 
using various kinds of efficient scene representation techniques.
Interestingly, not only are these approaches faster, they also tend to yield better image synthesis results. Other works have focused more directly on improving the quality of the synthesis results~\citep{Zhang20f,Barron21,Barron22,Barron23,Turki24}, 
by introducing unbounded scene representations or reducing aliasing artifacts. Nerfacto~\citep{Tancik23}, introduced in the popular Nerfstudio project, combines many components of these approaches into an integrated one. Hence, we use it as the basis for implementing our own approach. 

\parag{Gaussian Splatting.}

3D Gaussian Splatting~\citep{Kerbl23} has recently emerged as a powerful alternative to NeRFs for novel view synthesis results using an explicit point-based representation. It is arguably better at novel-view synthesis but NeRF-based models continue to shine in many scenarios, offering unique advantages through their continuous representations. This is the case in applications such as surface reconstruction~\citep{Li23c}, 3D scene understanding~\citep{Kerr23} and inverse rendering~\citep{Jin23}. In addition, the general regularization method we propose could be potentially used in Gaussian Splatting as well, by simply changing the point sampling method to a Gaussian placement one.

\parag{Regularizers.} 

A straightforward approach to improving the performance of NeRFs is to incorporate geometric priors to regularize and guide the training process. To this end, many methods have been proposed. They rely on depth guidance~\citep{Deng22c, Wang23a, Wang23b, Roessle22, Yu22}, geometric constraints~\citep{Niemeyer22, Somraj23, Truong23, Kim22b}, or pre-training on similar scenes~\citep{Chen21f, Yu21c, Wang21i, Wu23, Xu23e}. However, these methods are all plagued by generalization issues. For depth priors, it is difficult to obtain accurate depth predictions, especially in real-world unbounded scenes. The geometry-based constraints often fail to properly refine the results in complex unbounded scenes. Prediction-based methods
are mostly restricted to a 3-view setting in bounded scenes, due to the limitations of a prediction-based architecture and the limited availability of real-world unbounded scene data with 3D ground truth. Note also that many methods have been proposed for only sparse-view cases~\citep{Jain21,Uy23,Yang23c,Somraj23,Somraj23b,Shi24,Zhu24b,Zhang24}, which however are restricted in denser-view settings. Recently, ReconFusion~\citep{Wu2024} has been proposed to incorporate diffusion prior into NeRF training. 
This method works well for indoor or bounded scenes, but for open-space scenes the performance drops drastically as the original paper shows. 

\parag{Image Features.} 

Recently, there has been tremendous progress in large-scale self-supervised pre-training using Masked image Modeling~\citep{He22a, Wei22b, Zhou21b} (MIM). These new techniques yield foundation models, such as DINOv2~\citep{Oquab23}, which encode local geometric information better than classification pretraining~\citep{Xie23}, which makes them very useful in a NeRF context. Image features that even better for image matching can be obtained~\citep{Yi16b,Sun21a,Edstedt23,Sarlin20,Tyszkiewic20,Truong23}. However the pairwise matching schemes they are mostly intended for is not the best one for handling large image collections because it does not leverage multi-image cues.

\section{Methodology} 

We now introduce our \textit{View-consistent Sampling} (VS-NeRF) approach for NeRF training. As our method is built on the standard NeRF framework, we first describe it briefly in \cref{sec:basics}. Next, we describe our approach to distill high-level image features and preserve only view-consistent information from the foundation model DINOv2~\citep{Oquab23} in \cref{sec:distillation}. We then introduce a sampling mechanism to exploit these features as well as color features along camera rays in \cref{sec:sampling}. Finally, we describe the proposed depth-pushing loss as a weaker regularization to force the model to favor distant samples in in \cref{sec:depth}. 

\subsection{NeRF Basics} \label{sec:basics}

\paragraph{Scene Representation.}

The 3D scene is generally represented by an MLP, and optionally and additional feature grid, which encode both geometry information and view-dependent color information. Specifically, the geometry of the scene is encoded by the neural network as a function $f: \mathbb{R}^3 \rightarrow \mathbb{R}$ that maps a spatial coordinate $\mathbf{x} \in \mathbb{R}^3$ to its corresponding volume density value $\sigma$. The view-dependent color information is encoded by the network as a function $f: \mathbb{R}^3 \times \mathbb{S}^2 \rightarrow \mathbb{R}^3$ that takes a point coordinate $\mathbf{x} \in \mathbb{R}^3$ as well as a viewing direction $\mathbf{d}$ as input and outputs the associated view-dependent color value $\mathbf{c} = (r, g, b)$.

\parag{Volume Rendering.} The rendering process is of critical importance because it associates a 3D representation of the scene with 2D images, which makes the use of image reconstruction loss possible. In the NeRF literature, the most frequently used rendering technique in 3D vision tasks is known to be volume rendering. Given a ray $\mathbf{r}(t) = \mathbf{o} + t \mathbf{d}$, the volume rendering equation yields the color of one pixel in the 2D image corresponding to the ray $\mathbf{r}$ by evaluating 
\begin{equation}\label{eq:volumerendering}
\mathbf{\hat{C}}(\mathbf{r}) = \int_{t_n}^{t_f} \omega(t) \mathbf{c}(\mathbf{r}(t), \mathbf{d}) dt \; ,
\end{equation}
where $\omega(t) = T(t)\sigma(\mathbf{r}(t))$ is the weight function, $\sigma$ represents the volume density, $\mathbf{c}$ represents the directional color, and $T(t) = \exp(-\int_{t_n}^{t} \sigma(\mathbf{r}(s)) ds)$ represents the transparency function.

In practice, this integral is evaluated by sampling the ray in discrete locations, which is of vital importance depending on the sampling scheme~\citep{Barron22}. NeRF volume rendering is then performed by accumulating color values from $S$ samples $(t_i)_{1 \leq i \leq S}$ along a ray $\mathbf{r}$. This yields

\begin{equation} \label{eq:volume_rendering}
    \mathbf{\hat{C}}(\mathbf{r}) = \sum_{i=1}^{S} T_i (1 - \mathrm{exp}(- \sigma_i \delta_i)) \mathbf{c}_i, \ T_i = \mathrm{exp}\left(- \sum_{j=1}^{i-1} \sigma_j \delta_j\right) \;
\end{equation}

where $\delta_i = t_{i+1} - t_{i}$ is the distance between two consecutive samples.

\parag{Loss Function.} 

Given the estimated color $\mathbf{\hat{C}}(\mathbf{r})$ of \cref{eq:volume_rendering}, let $\mathbf{C}(\mathbf{r})$ be the corresponding pixel true color. Using these notations, we can define an MSE loss 
\begin{equation}\label{eq:color_loss}
\mathcal{L}_{color} = \frac{1}{|\mathcal{B}|} \sum_{\mathbf{r} \in \mathcal{B}}  \| \mathbf{\hat{C}}(\mathbf{r}) - \mathbf{C}(\mathbf{r}) \|^2 \; ,
\end{equation}
where $\mathcal{B}$ denotes a randomly chosen batch of rays and $|\mathcal{B}|$ denotes the batch size. The weights of the NeRF scene representation network are computed by minimizing this loss, using a different batch of rays at each iteration.  

\begin{figure*}[t]
	\setlength\mytmplen{.200\linewidth}
	\setlength{\tabcolsep}{1pt}
	\renewcommand{\arraystretch}{0.5}
	\centering
	\begin{tabular}{ccccc}
        Test Image A & Test Image B & DINOv2 & Distilled  \vspace{0.1em} \\
		\includegraphics[height=\mytmplen]{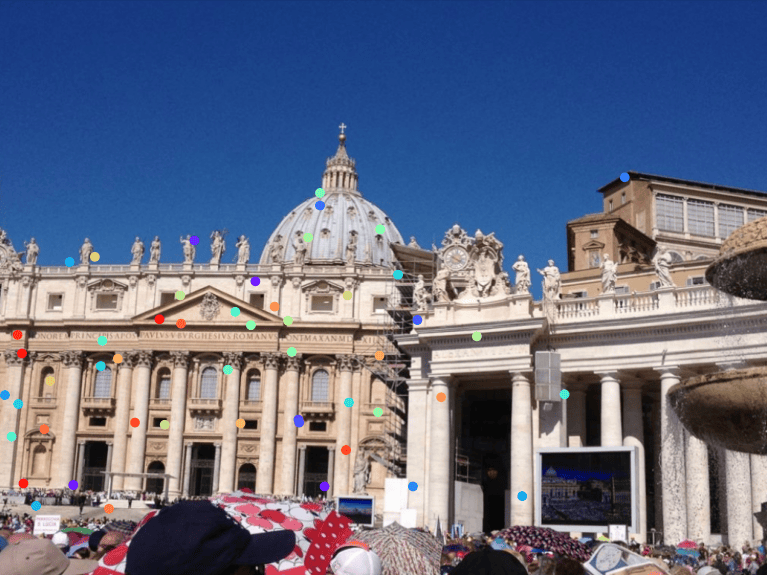} &
        \includegraphics[height=\mytmplen]{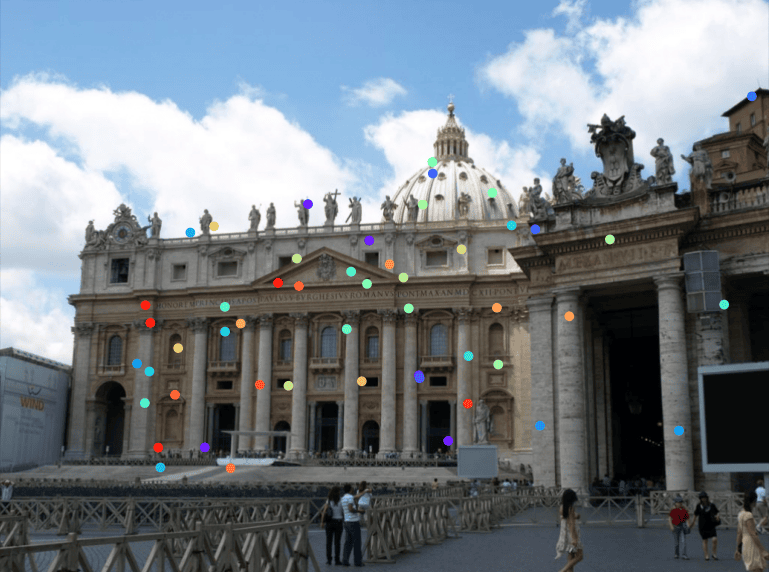} &
        \includegraphics[height=\mytmplen]{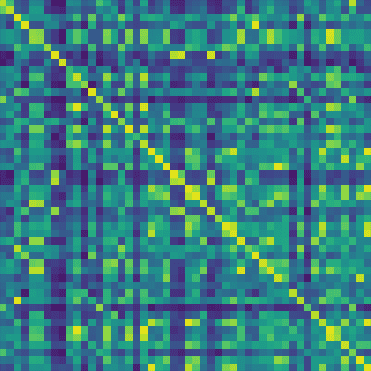} &
        \includegraphics[height=\mytmplen]{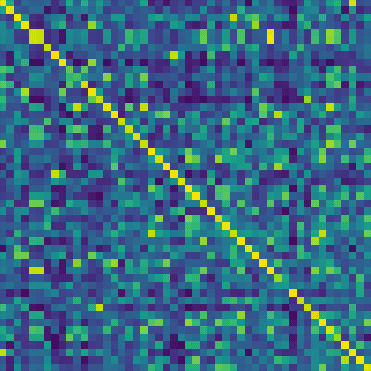} &
        \includegraphics[height=\mytmplen]{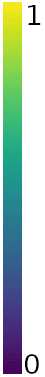} \\
    \end{tabular}
    \vspace{-.1cm} %
	\caption{
		\label{fig:distillation}
		Visualization of the feature distillation process. For the two test images from Megadepth dataset~\citep{Li18o}, we first randomly generate $50$ ground truth correspondences (same as in the training process), shown as colored dots, and then extract vanilla DINOv2 features ($384$ dimension) and the proposed distilled DINOv2 features ($32$ dimension) at these locations.
		We compute the feature similarities across the two views and show the resulting similarity matrices on the right, where an optimal correspondence should give the identity matrix.
	}
\end{figure*}

\subsection{Distillation of Geometric Information} \label{sec:distillation}

To form meaningful view-consistent statistics for adaptive sampling, a good representation of images in the context of multi-view captures is critical.
In this paper, we propose to use foundation models that provide powerful general-purpose visual features, e.g. DINOv2~\citep{Oquab23}, for extracting such high-level representations containing crucial context information. Note that there are alternatives to DINOv2 such as features from diffusion models~\citep{Luo24}, which we found to deliver similar results. Since diffusion models are slower in inference time, we opted to use DINOv2. However, there are two main hurdles in utilizing the features from foundation models. Firstly, as general-purpose features, the output of a foundation model encodes image information in many different aspects that are useful in different tasks. In the NeRF setting, we are particularly interested in geometric information that can be expected to be similar across multi-view images of the same scene. Secondly, the dimensionality of features from foundation models is in general very high, e.g. $384$ in DINOv2, resulting in prohibitively large memory consumption in sampling.

In this paper, we propose to resolve the issues by distilling geometric information from the foundational features. Note that we use the term {\it distillation} in a different way than in \textit{Feature Field Distillation} papers, such as~\citep{kobayashi22}, which focuses on lifting 2D features to a 3D representation. We distill features by extracting geometric information from redundant high-dimensional image features. Inspired by~\citep{Luo24}, we use a very lightweight Resnet bottleneck block~\citep{He16a} to project the high-dimensional features to a lower dimension for distillation. To supervise the distillation process, we adopt the Megadepth dataset~\citep{Li18o} which provides 3D ground truth and is prevalently used in geometric matching tasks. 

\paragraph{Training of Distillation Process.}

We adopt a very simple strategy for training. Specifically, in the training phase, we freeze the foundation model and only update the parameters in the Resnet bottleneck block. This leads to a much smaller number of training parameters and also requires much less data. We randomly choose $50000$ pairs of images from Megadepth to train. For each image pair, we use the ground truth depth map to randomly generate $50$ corresponding points, and extract the distilled features at the point locations on the image pairs. 
We then supervise the network using a symmetric cross entropy loss, in the same fashion as CLIP~\citep{Radford21}, to make extracted features in corresponding locations as close as possible while still being distinctive from other features. A visualization of the distillation process can be found in \cref{fig:distillation}. As a result, in our experiments we can reduce the feature dimensionality by a factor of around $10$, e.g. $384$ from DINOv2 to $32$, without compromising on useful geometric information. Please refer to the ablation studies in \cref{sec:ablations} for the discussion of optimal dimensionality in NeRF settings.
 
\begin{figure*}[t]
	\setlength\mytmplen{.132\linewidth}
	\setlength{\tabcolsep}{1pt}
	\renewcommand{\arraystretch}{0.5}
	\centering
	\begin{tabular}{ccccc}
        Reference Point & VC Distribution & Reprojection A & Reprojection B & Reprojection C \vspace{0.2em} \\ 
		\includegraphics[height=\mytmplen]{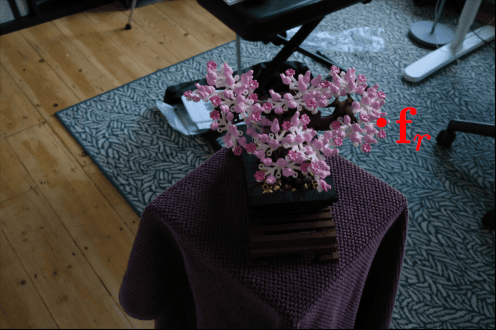} &
        \includegraphics[height=\mytmplen]{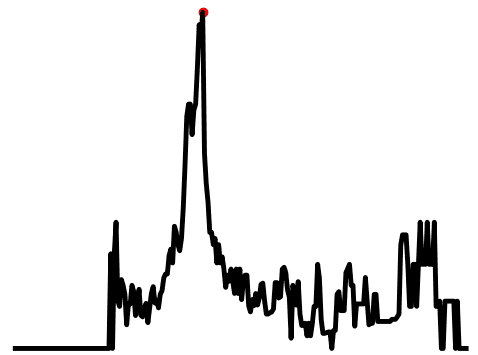} &
        \includegraphics[height=\mytmplen]{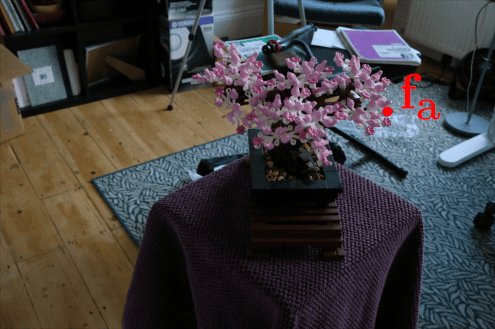} &
		\includegraphics[height=\mytmplen]{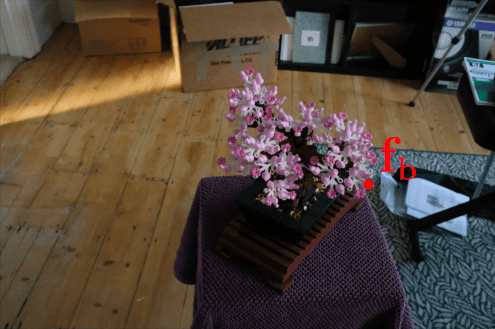} &
        \includegraphics[height=\mytmplen]{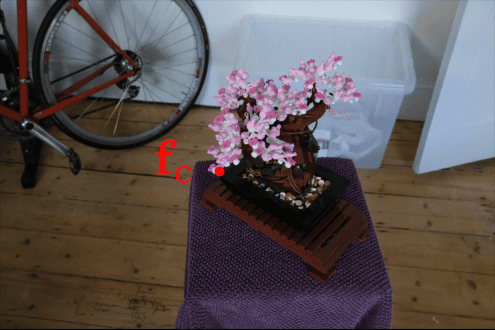} \\
    \end{tabular}
    \vspace{-.1cm} %
	\caption{
		\label{fig:vc_distribution}
		Visualization of the effectiveness of the view-consistency metric on the \textsc{Bonsai} scene from the MipNeRF360 dataset. As shown, we shoot a ray from the reference point in the leftmost image, compute the view-consistency metric distribution along the ray, and reproject the peak point in the distribution onto other views. The projections of the peak are consistent and correspond to a surface point.
	}
\end{figure*}

\subsection{View-Consistent Sampling} \label{sec:sampling}

In the original NeRF and most NeRF variants, the volume rendering of \cref{eq:volume_rendering} is typically achieved using naive sampling strategies such as uniform, stratified, or linear disparity sampling.
Hence, there is no prior in the sampling process and hence no regularization while learning the radiance and density fields. When there are abundant input views, this is usually not an issue but it can result in unwanted artifacts with a smaller set of input images.

VS-NeRF remedies this by making the sampling adaptive based on a prior: it samples more densely the 3D locations whose projections have view-consistent features because they are more likely to correspond to 3D surface points. This requires both a view-consistency metric and an adaptive sampling scheme based on that metric, which we describe in Sections \ref{sec:vcmetric} and \ref{sec:adap_sampling}, respectively. A graphical visualization of the proposed view-consistent sampling technique can be seen in \cref{fig:keyidea}.

\paragraph{Sampling Setup.}
We assume that we have a collection of $N$ posed images $\{\mathbf{I}_i\}_{i=1}^{N}$, from which we generate image feature representations $\{\mathbf{F}_i\}_{i=1}^{N}$. As in NeRF and most of its variants, 
the training is performed by repeatedly sampling a batch of rays. For each ray $\mathbf{r}$, we initially need to place pre-samples along the ray $(t_i)^{pre}_{1 \leq i \leq M}$, to obain $M$ points $\{\mathbf{p}_i\}_{i=1}^{M}$. Note that these pre-samples are different from the initial samples in NeRF, as pre-samples are for computing view-consistency statistics only and will not be used to compute losses.
The pre-samples are generated uniformly within a distance, but after a fixed threshold the step sizes will increase with each sample due to scene contraction. This strategy is the same as the initial sampling strategy in Nerfacto~\citep{Tancik23} and we refer to the original paper for details. 

\subsubsection{Computation of View-Consistency Metric}\label{sec:vcmetric}

\paragraph{Features from Projections.} As shown in \cref{fig:vc_distribution}, the feature representation at the pixel location where the ray comes from is denoted as reference feature $\mathbf{f}_r$. For each pre-sample point $\mathbf{p}_i$, we can project it onto an arbitrary view $v_j$. If the point $\mathbf{p}_i$ is visible to $v_j$, then naturally by interpolating over the feature representation $\mathbf{F}_j$, we can 
obtain the projection feature $\mathbf{f}_{ij}$. Due to limited field-of-view (FOV) of cameras, there is a varying number of views that a point $\mathbf{p}_i$ can be projected onto. We denote the set of views that a point $\mathbf{p}_i$ can be projected onto as $V_i$, and 
$|V_i|$ as its cardinality. 

\paragraph{Normalized Similarity Measure.} In this paper, we jointly use high-level distilled features, and plain normalized RGB values as low-level color features.
Please see the ablation study in \cref{sec:ablations} to understand their respective effects. However, the two kinds of features are defined in different metric spaces. That is to say, while Euclidean distances can be used for measuring discrepancies among color features, cosine similarities are most frequently used as a distance measure of features from pre-trained models such as DINOv2. 
In this paper, we use a normalizing strategy to convert the measures among features to binary numbers, be it color feature or distilled feature. In particular, along an arbitrary ray $\mathbf{r}$, we first compute the measures between
the reference feature and projection features from all sampled points $\{m(\mathbf{f}_r, \mathbf{f}_{ij}) \  | \ j \in V_i$ \}, be it Euclidean distances or cosine similarities. We then normalize the set of measures $m$ based on its variance $\sigma_m = \sqrt{\frac{1}{N_m} \sum_{i=1}^{N_m} m_i^2} $, and take the negative if the measure
is Euclidean distance to align distance with similarity. Thus, we have defined the normalized similarity measure $\{m_n(\mathbf{f}_r, \mathbf{f}_{ij}) \  | \ j \in V_i$ \}.

\paragraph{View-consistency Metric.}

Given the normalized similarity measure $\{m_n(\mathbf{f}_r, \mathbf{f}_{ij}) \  | \ j \in V_i$ \}, we assume its values follow a normal distribution and we experimentally determine a reasonable threshold $\delta$ accordingly. The view-consistency metric of point $\mathbf{p}_i$ along the ray is computed as:

\begin{equation} \label{eq:vc_metric}
    s_i = \frac{1}{|V_i|} \sum_{j \in V_i} \mathds{1} \{m_n(\mathbf{f}_r^c, \mathbf{f}_{ij}^c) > \delta \ \land m_n(\mathbf{f}_r^d, \mathbf{f}_{ij}^d) > \delta\} \; ,
\end{equation}
where $\mathds{1}$ denotes the indicator function, superscripts $c$ and $d$ denote \textit{color} and \textit{distilled} in projection features respectively. Intuitively, \cref{eq:vc_metric} measures the average view consistency over the views the point can be projected onto. 
Although occlusions may hinder the effectiveness of this metric, statistically the score is still prominent for surface points. A visualization of the computed view-consistency metrics along a ray using real data can be seen in \cref{fig:vc_distribution}. 

\begin{figure*}[t]
	\setlength\mytmplen{.245\linewidth}
	\setlength{\tabcolsep}{1pt}
	\renewcommand{\arraystretch}{0.5}
	\centering
	\begin{tabular}{cccc}
        Ground Truth & Nerfacto & +Multi-view Depth & +Ours \\
		\includegraphics[width=\mytmplen]{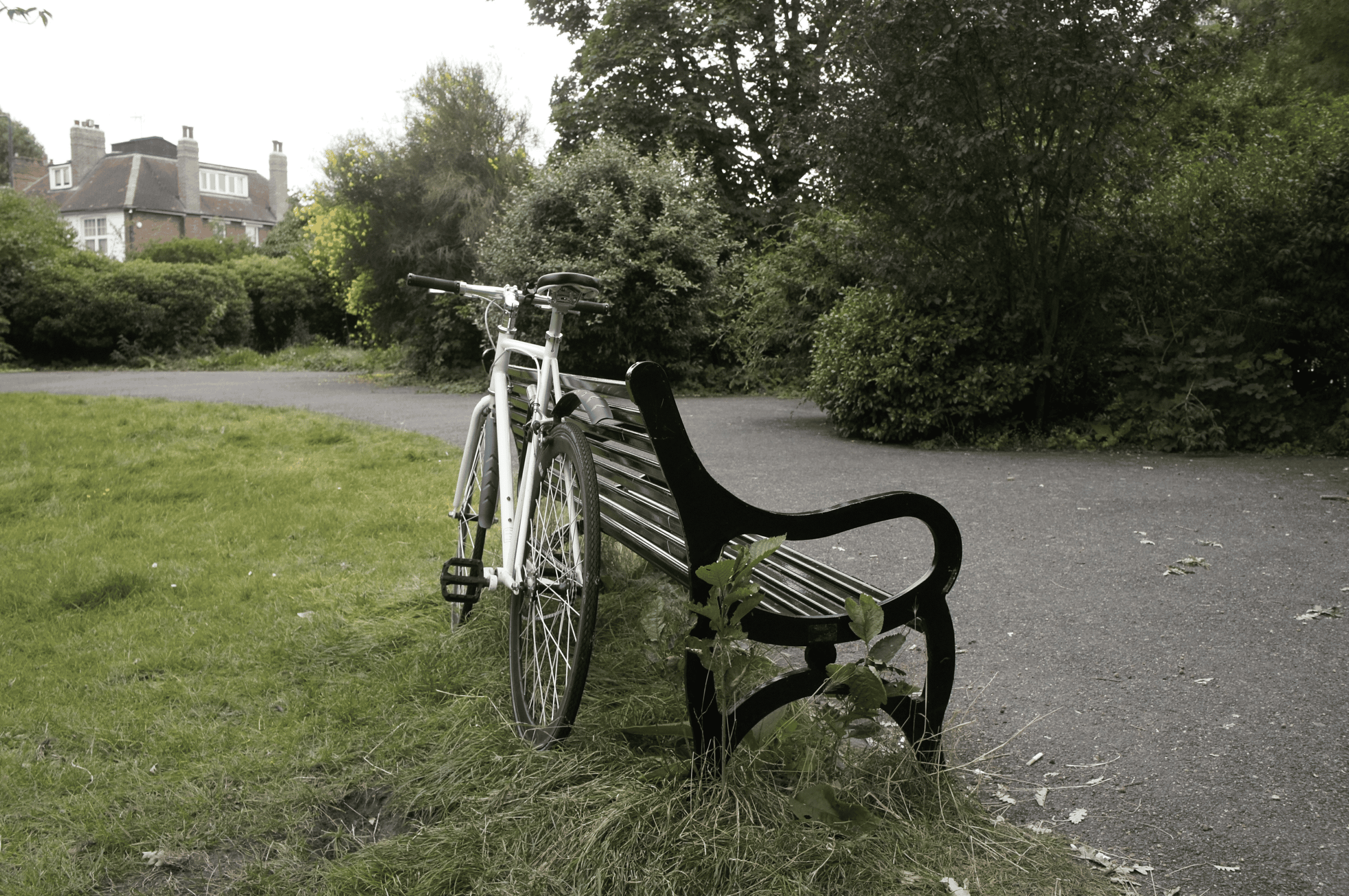} &
        \includegraphics[width=\mytmplen]{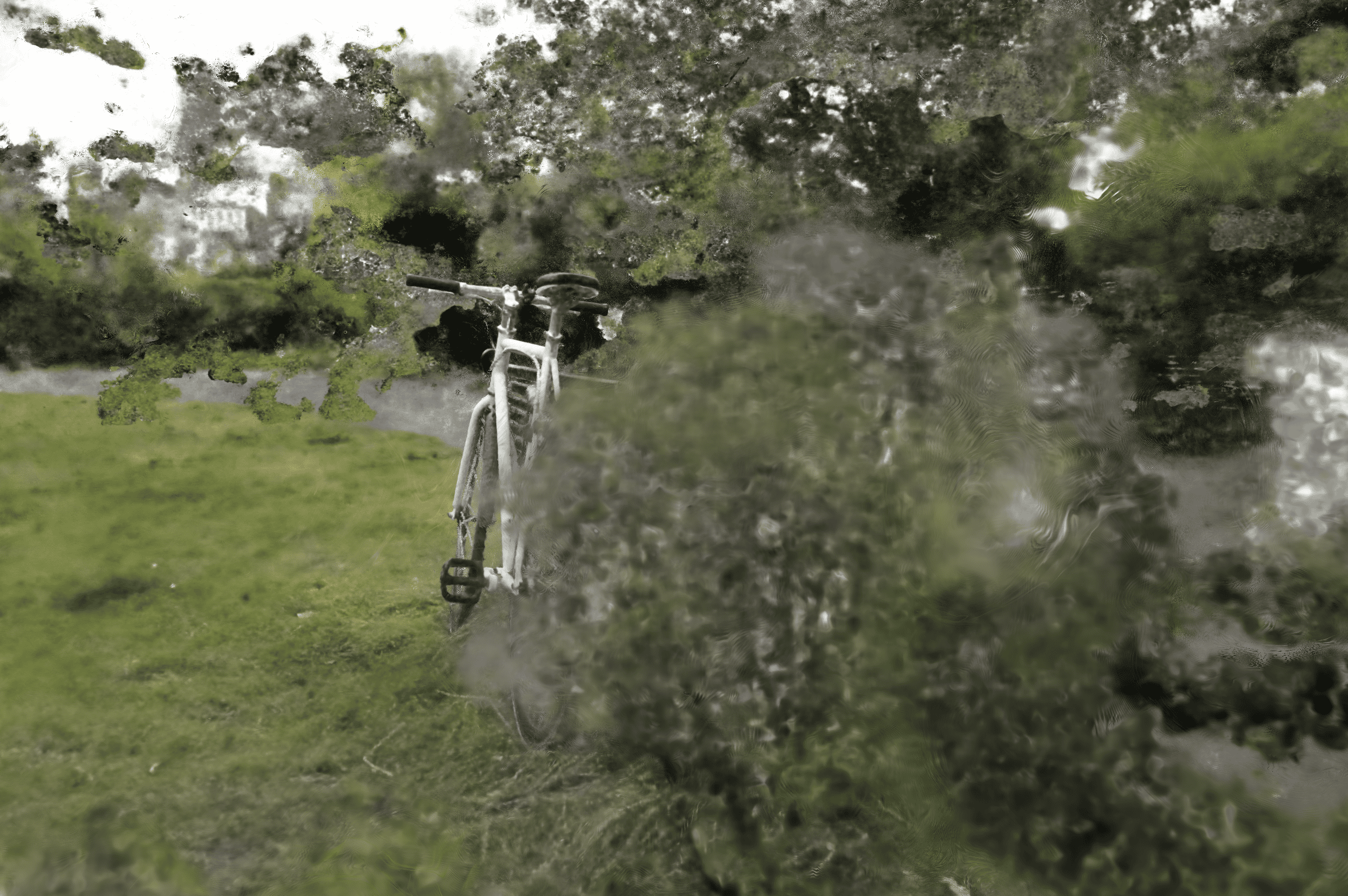} &
        \includegraphics[width=\mytmplen]{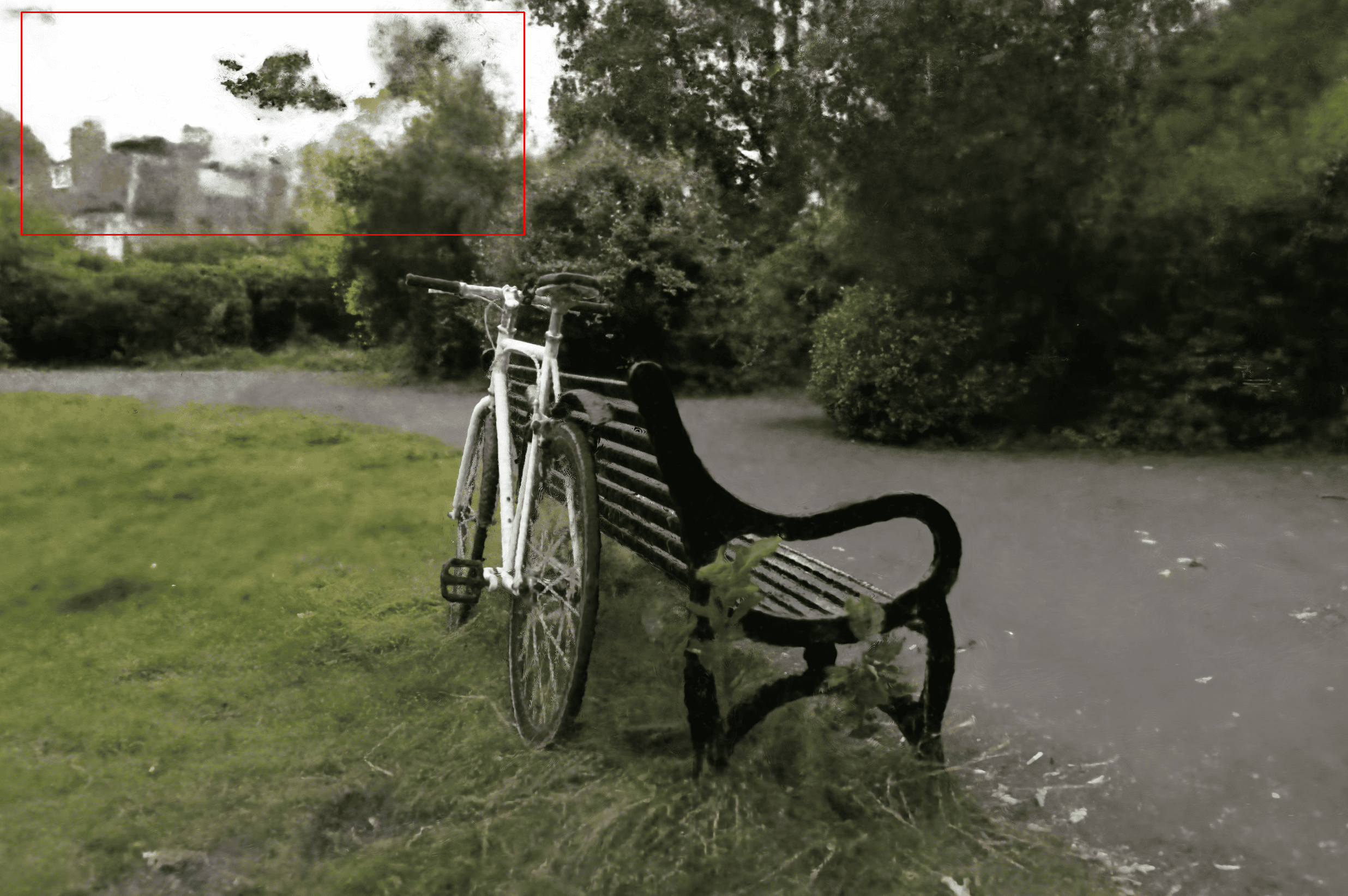} &
        \includegraphics[width=\mytmplen]{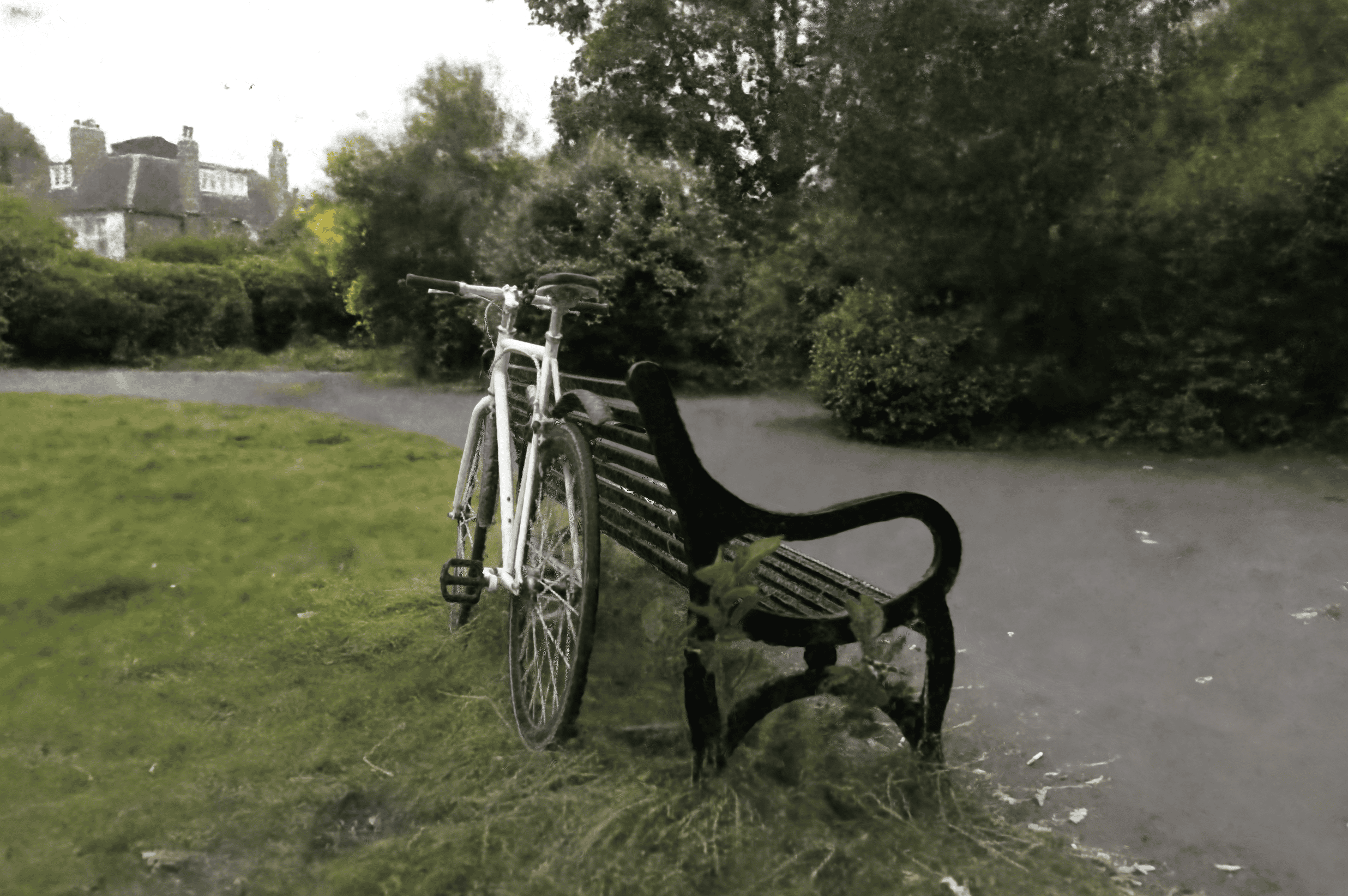} \\
        \includegraphics[width=\mytmplen]{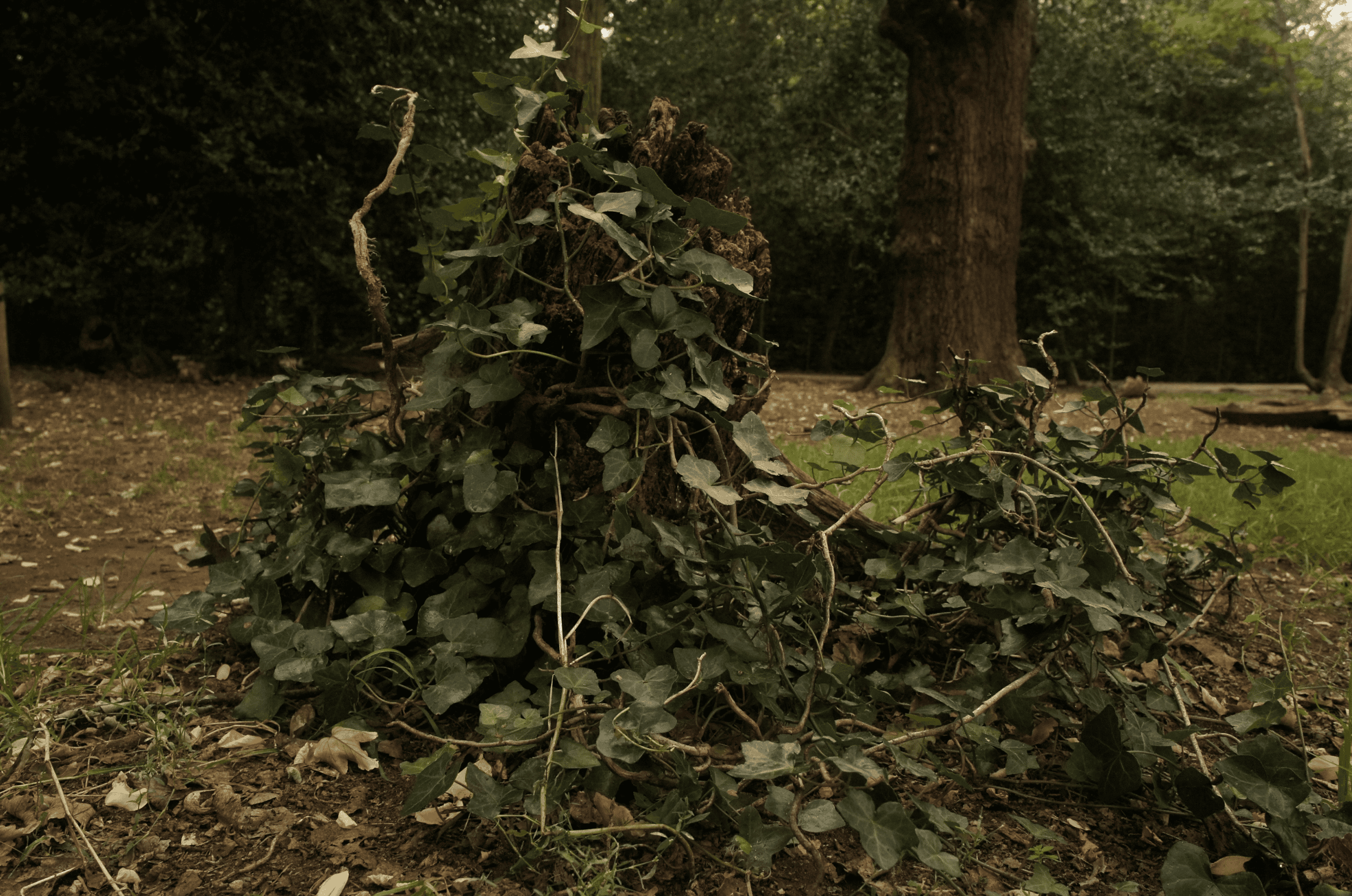} &
        \includegraphics[width=\mytmplen]{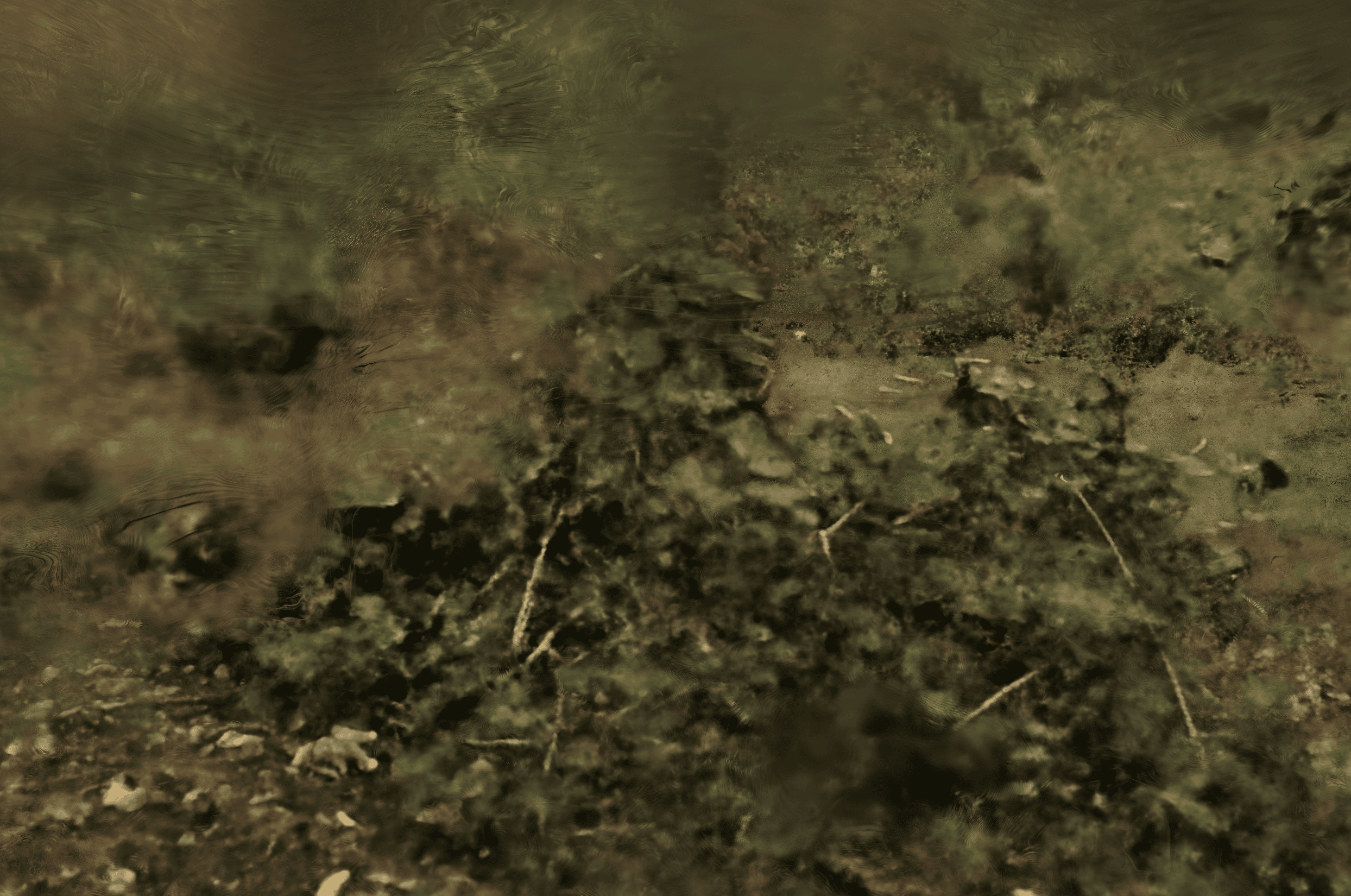} &
        \includegraphics[width=\mytmplen]{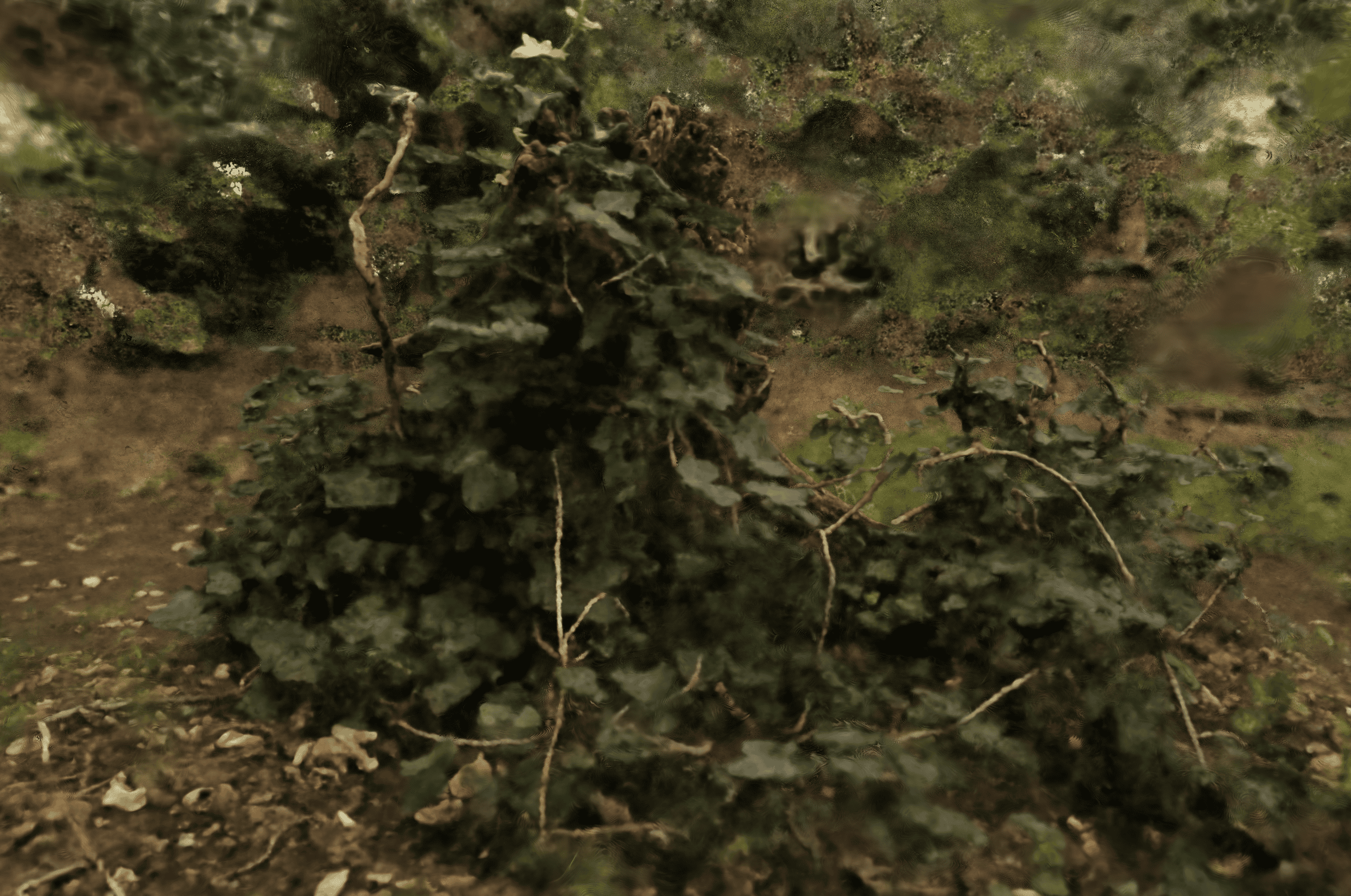} &
        \includegraphics[width=\mytmplen]{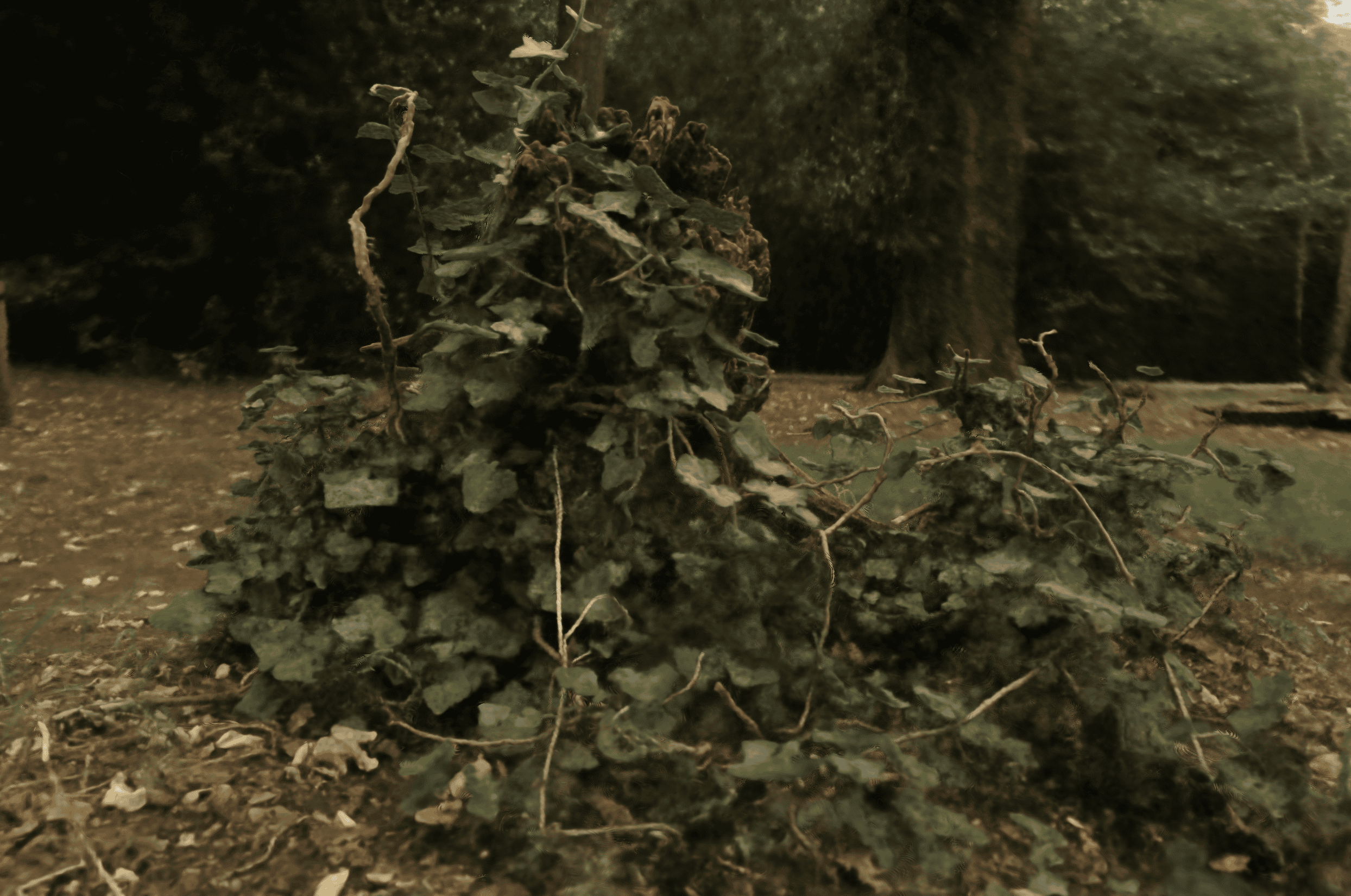} \\
        \includegraphics[width=\mytmplen]{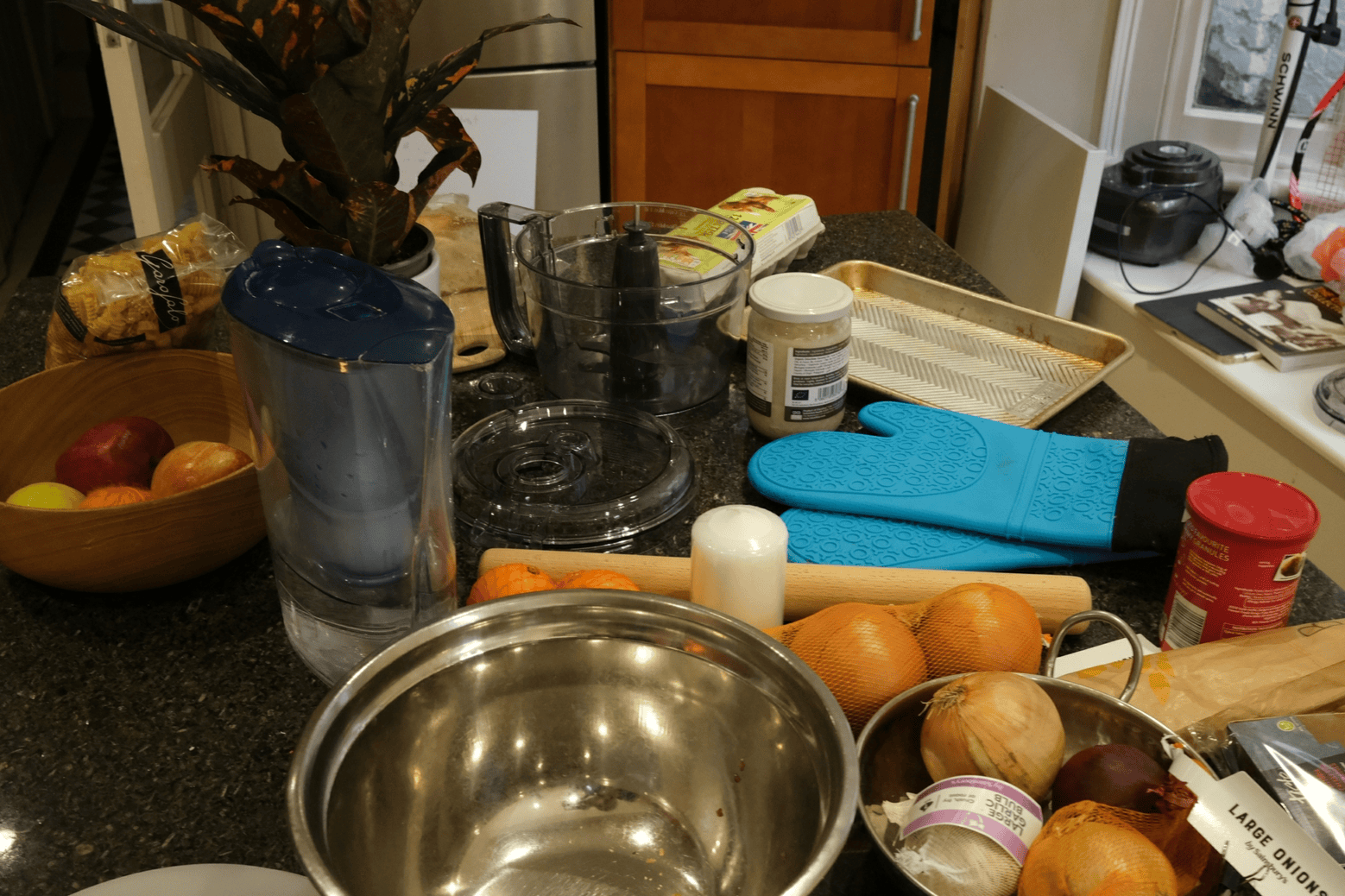} &
        \includegraphics[width=\mytmplen]{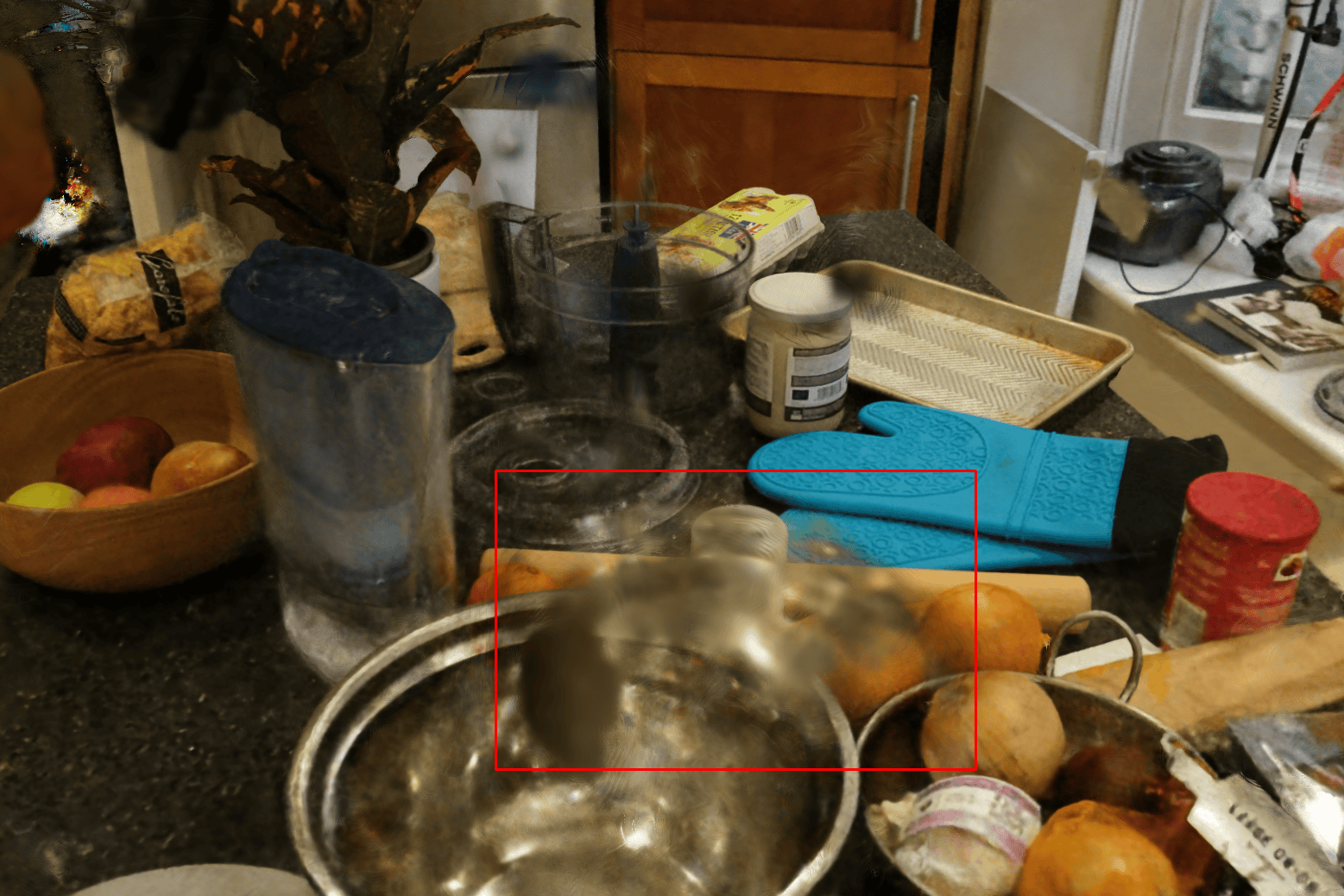} &
        \includegraphics[width=\mytmplen]{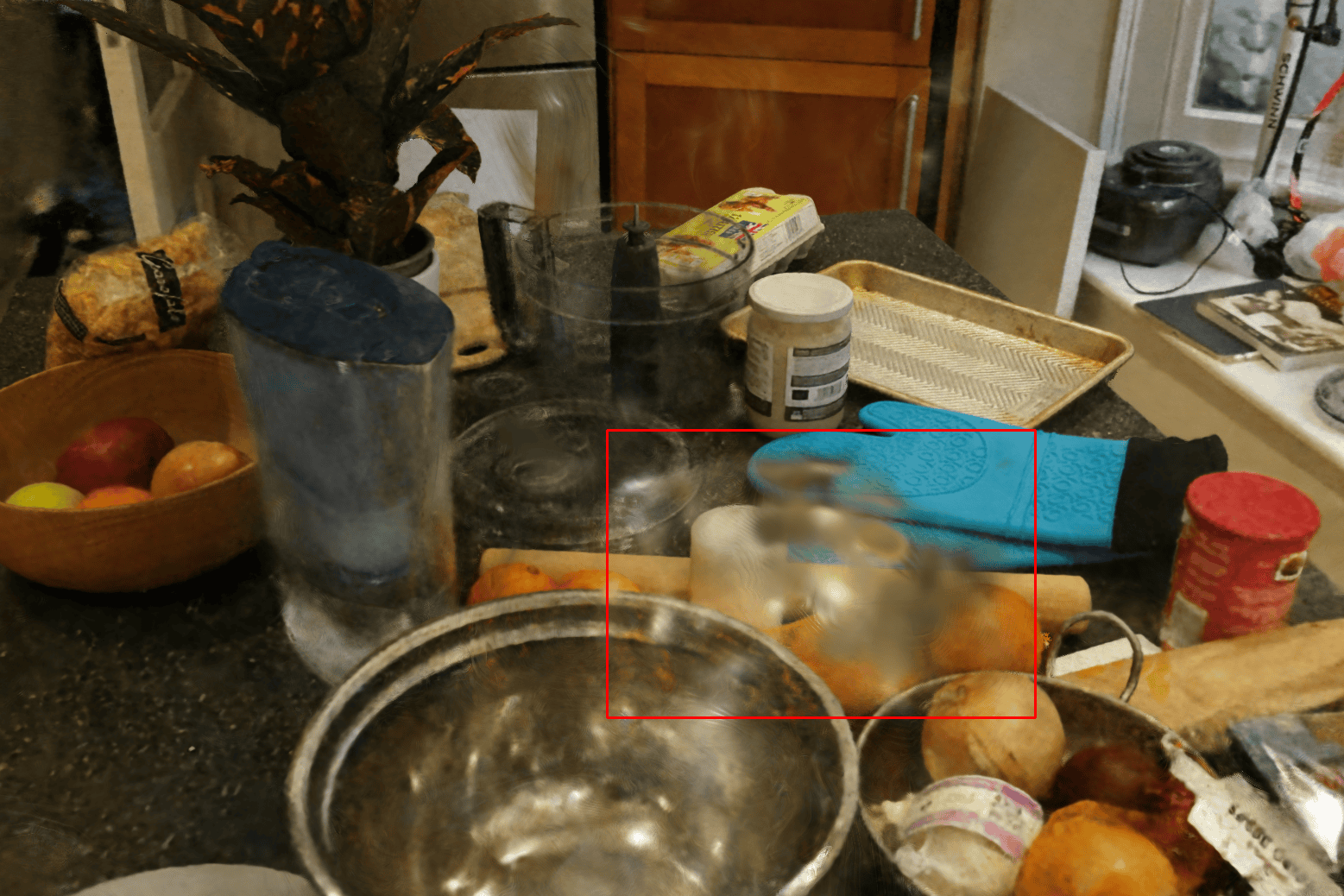} &
        \includegraphics[width=\mytmplen]{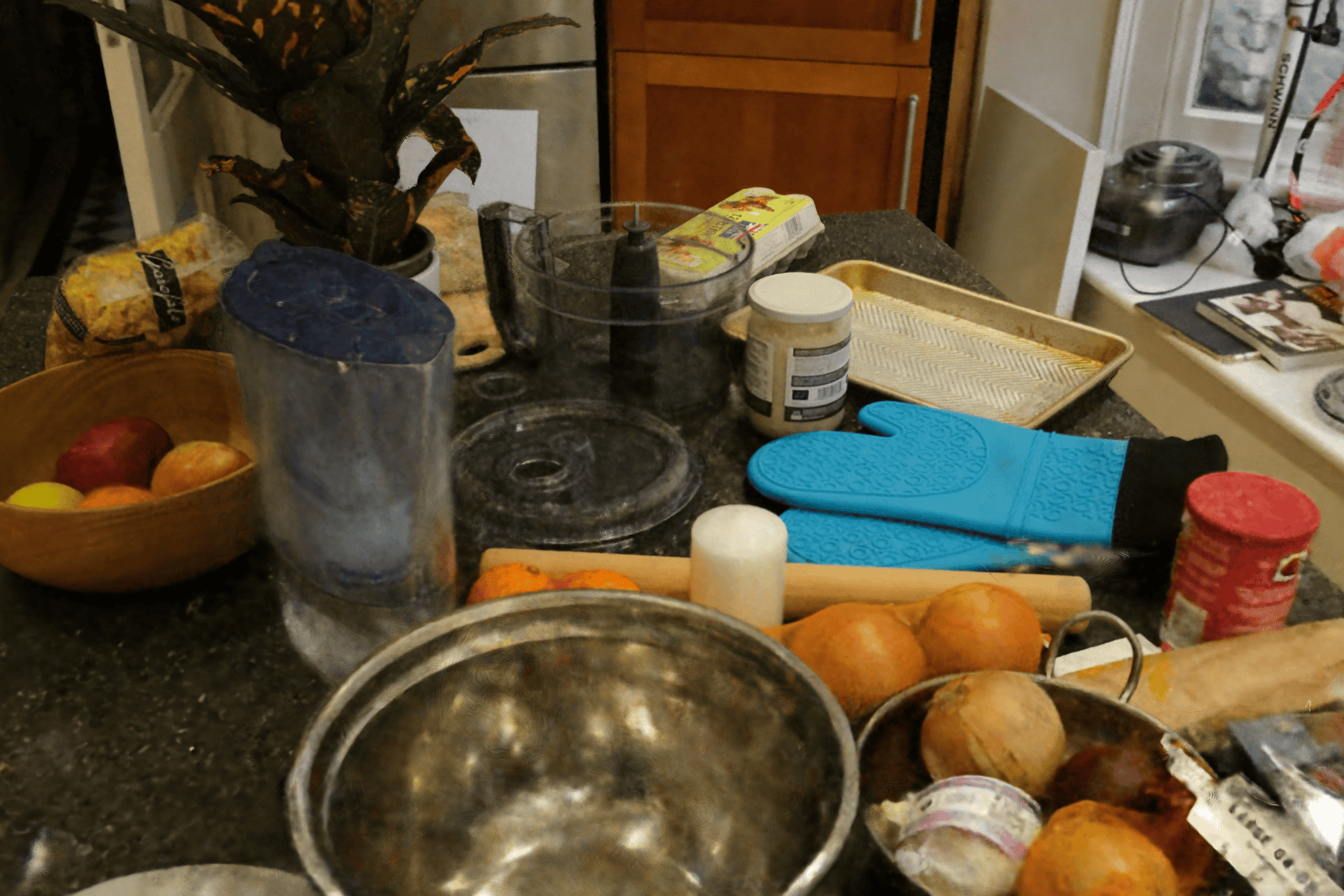} \\
        \includegraphics[width=\mytmplen]{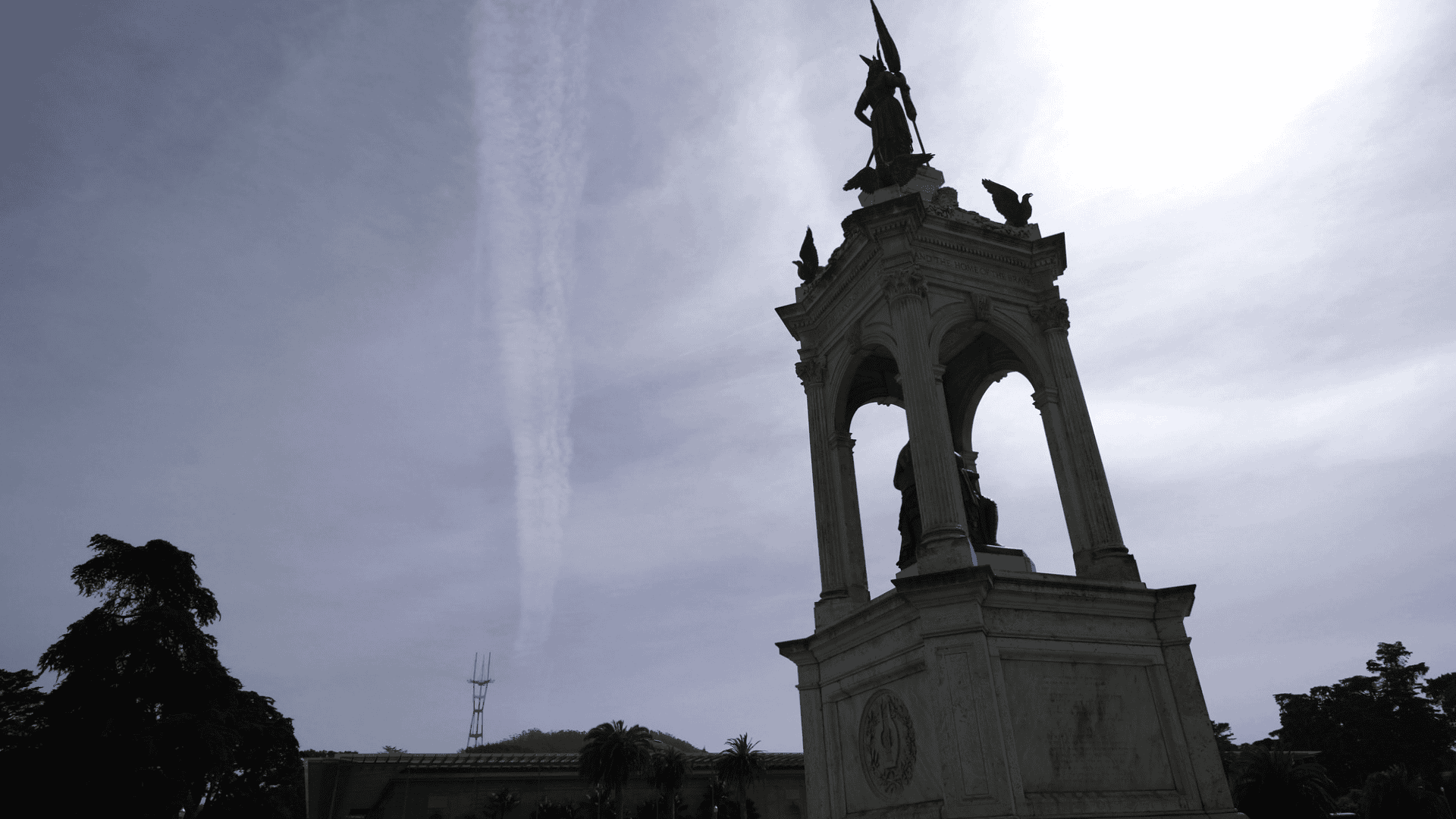} &
        \includegraphics[width=\mytmplen]{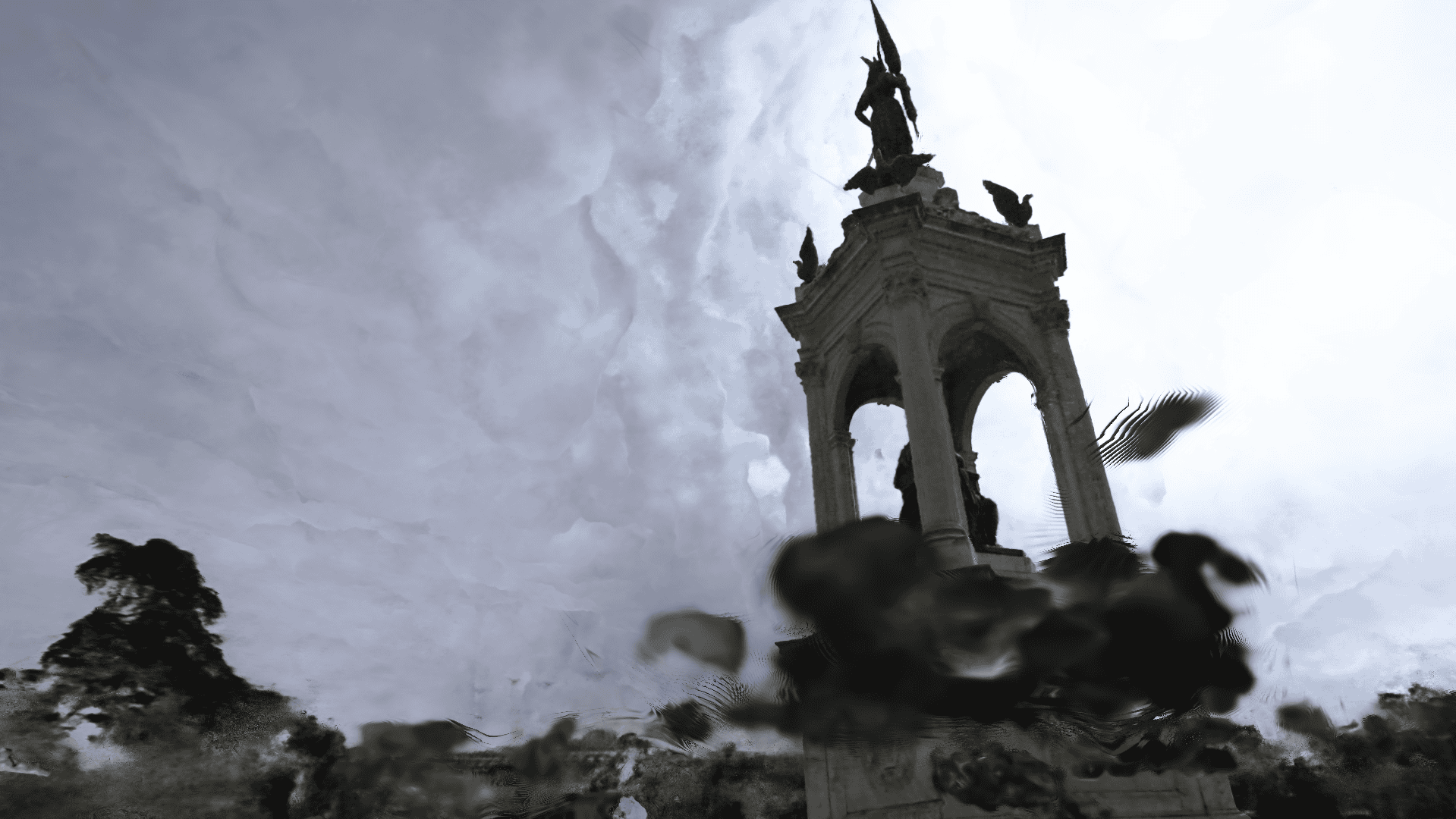} &
        \includegraphics[width=\mytmplen]{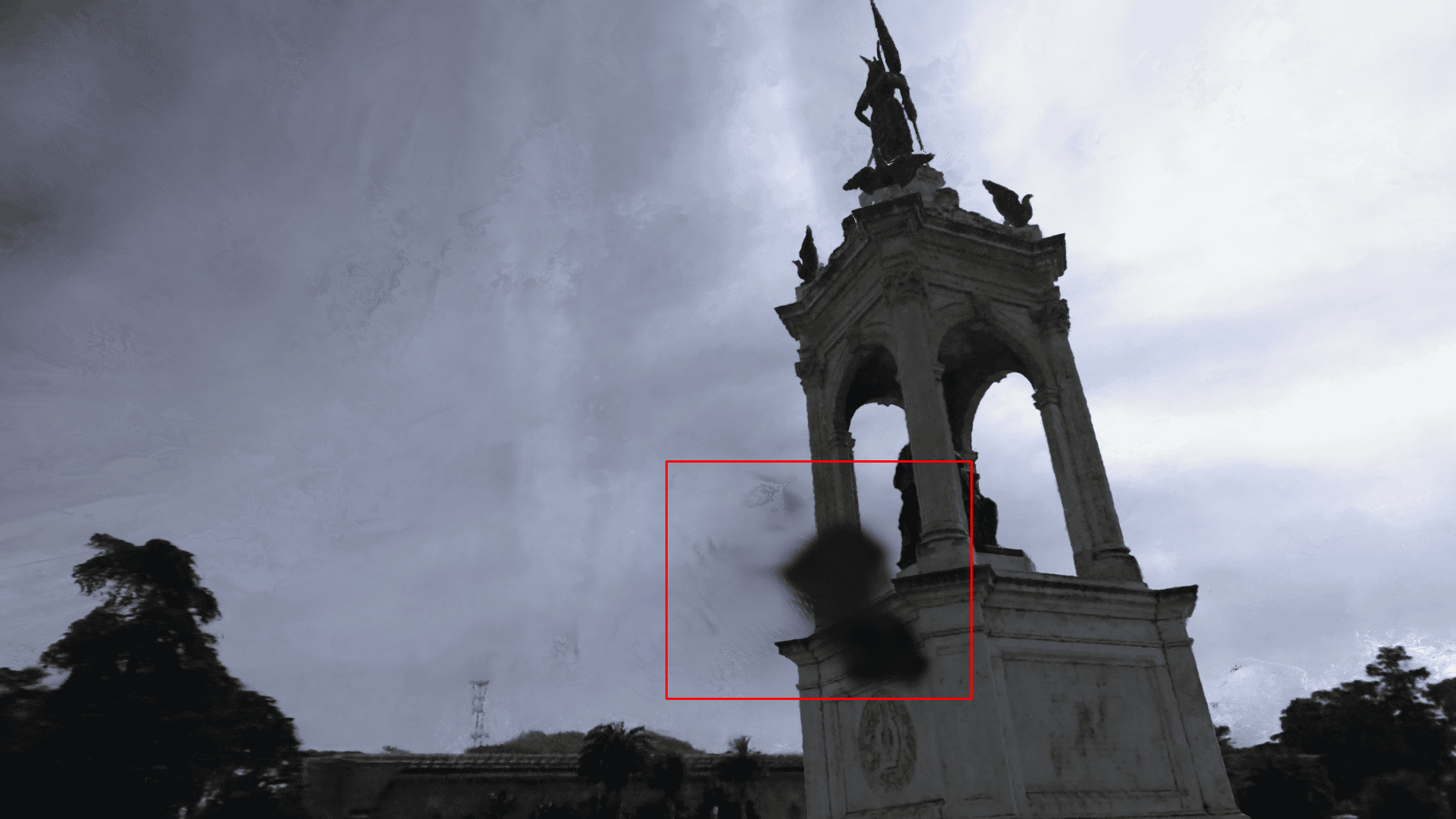} &
        \includegraphics[width=\mytmplen]{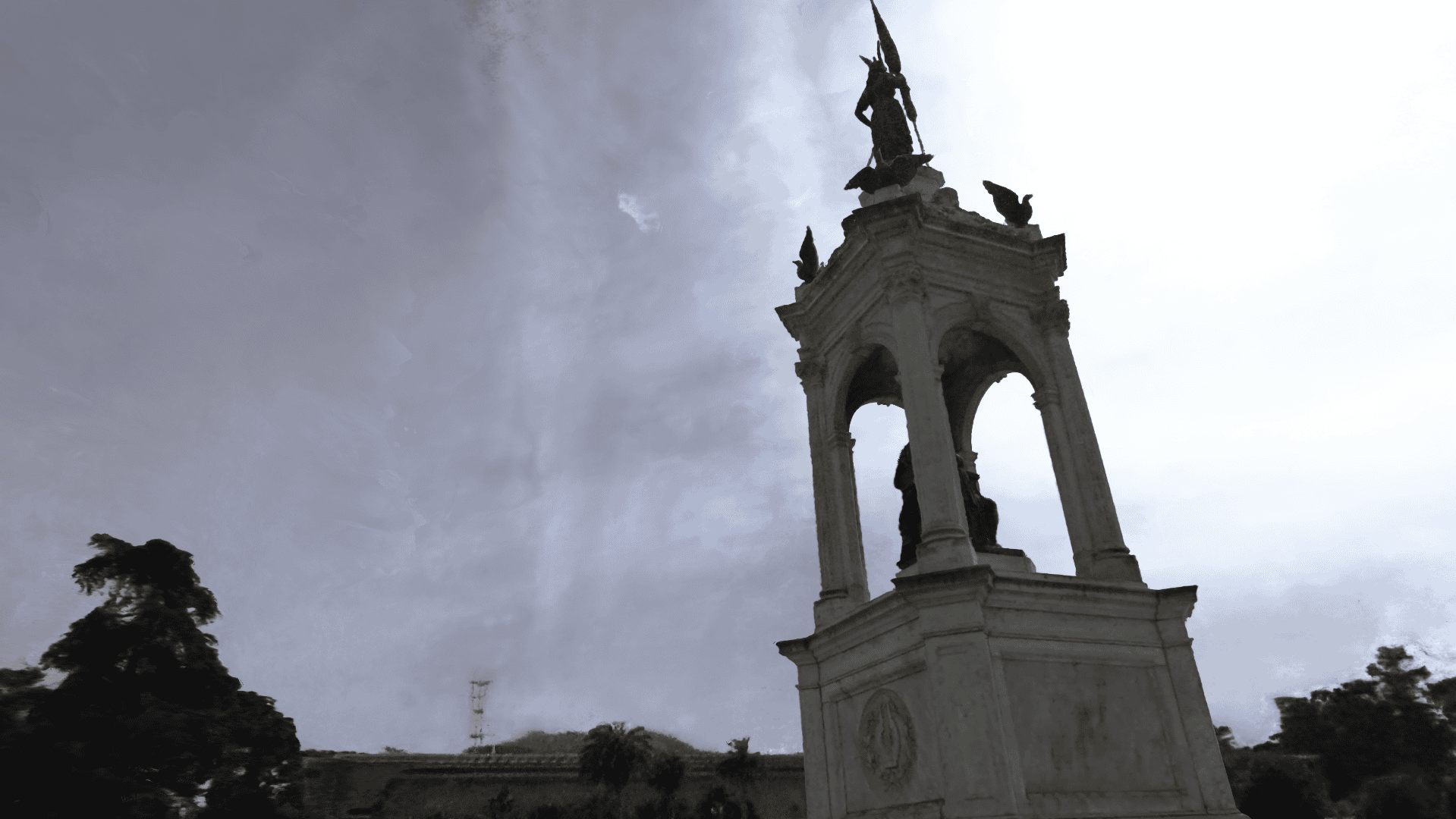} \\
    \end{tabular}
    \vspace{-.1cm} %
	\caption{
		\label{fig:comparisons}
		We show comparisons of VS-NeRF to the main competitors and the corresponding ground truth images from held-out test views. The scenes are, from the top down:
		\textsc{Bicycle} with $60$ training views, \textsc{Stump} with $110$ training views, \textsc{Counter} with $70$ training views from the Mip-NeRF360 dataset and \textsc{Francis} with $70$ training views from Tanks\&Temples. The '+' prefix indicates the included additional component to Nerfacto.
	}
\end{figure*}

\begin{figure*}[t]
	\setlength\mytmplen{.40\linewidth}
	\setlength{\tabcolsep}{1pt}
	\renewcommand{\arraystretch}{0.5}
	\centering
	\begin{tabular}{cc}
		\includegraphics[width=\mytmplen]{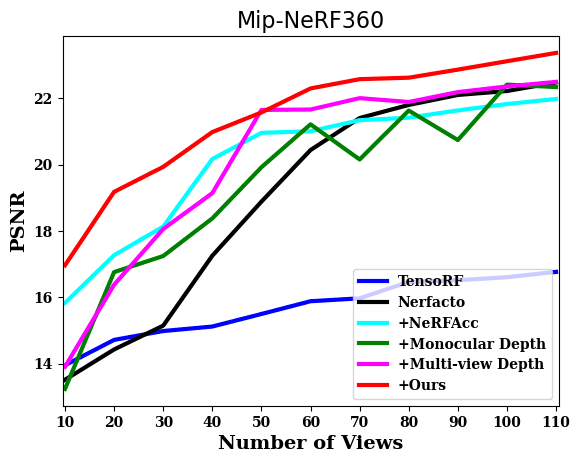} &
        \includegraphics[width=\mytmplen]{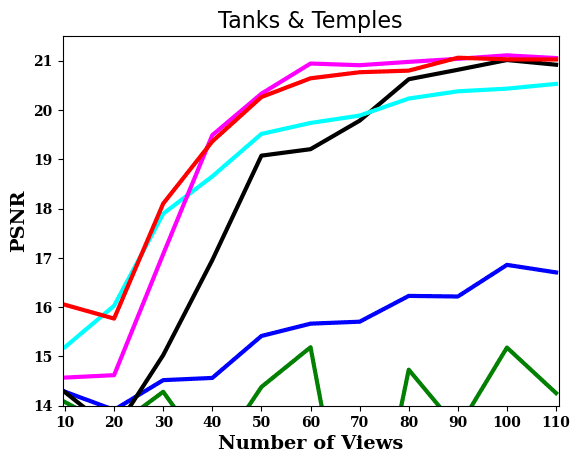} \\
    \end{tabular}
    \vspace{-.1cm} %
	\caption{
		\label{fig:curve_comparisons}
		We show performances of VS-NeRF and competitors with increasing number of views over Mip-NeRF360 dataset and Tanks\&Temples, in terms of PSNR values. The '+' prefix indicates the included additional component to Nerfacto.
	}
\end{figure*}

\subsubsection{Adaptive Sampling Scheme}\label{sec:adap_sampling}

Given the view-consistency metric of \cref{eq:vc_metric}, implementing view-consistent sampling becomes straightforward. After computing the metric for each pre-sampled point along the ray, we perform importance sampling based on the distribution along the ray.
Our rationale is that this view-consistentcy distribution is concentrated around the surface point, thus importance sampling from the distribution is the logical way to improve it.

In our implementation, we use the Probability Distribution Function (PDF) sampler from Nerfstudio~\citep{Tancik23} to perform importance sampling, which generatse samples that match a distribution. Specifically, as illustrated by \cref{fig:keyidea}, we first
compute view-consistency metrics from pre-samples $(t_i)^{pre}_{1 \leq i \leq M}$ along the ray. The PDF sampler will probabilistically sample the bins between consecutive pre-sample points, such that the distribution of number of samples in
each bin will match the view-consistency distribution, which gives the true samples $(t_i)_{1 \leq i \leq S}$ to compute losses as in \cref{eq:volume_rendering}.

\subsection{Depth-pushing Loss} \label{sec:depth}


In NeRF, the background of the scene is generally harder to reconstruct than foreground objects, typically because parts of the background may only be seen in very few views. To complement our adaptive sampling scheme, we introduce a depth-pushing loss which favors distant samples along the ray and provides a regularization to prevent background collapse. This shares a similar idea to prioritizing sampling distant points~\citep{Niemeyer22,Yang23c}. We write

\begin{align} \label{eq:depth_pushing_loss}
    \mathcal{L}_{depu} &= - \frac{1}{|\mathcal{B}|} \sum_{\mathbf{r} \in \mathcal{B}} \  \log(d(\mathbf{r}) + \varepsilon) \\
	\mathrm{~ where ~} d(\mathbf{r}) &= \sum_{i=1}^{S} T_i (1 - \mathrm{exp}(- \sigma_i \delta_i)) t_i \; ,
\end{align}
where $\varepsilon$ is a small constant that stabilizes the logarithm function near $0$ and $d(\mathbf{r})$ is the expected depth along the ray.  Its simple form makes it easy to integrate into the NeRF framework by adding it to the color loss of \cref{eq:color_loss}.


\renewcommand{\arraystretch}{1.0}
\begin{table*}[t]
    \small
	\centering
	\scalebox{1.15}{
		\begin{tabular}{l|cccc|cccc}
			\hline
			Dataset & \multicolumn{4}{c|}{Mip-NeRF360}  & \multicolumn{4}{c}{Tanks\&Temples} \\
			Method / Metric
			& \!PSNR$\uparrow$\!   & \!SSIM$\uparrow$\!    & \!LPIPS$\downarrow$\!  & Train
			& \!PSNR$\uparrow$\!   & \!SSIM$\uparrow$\!   & \!LPIPS$\downarrow$\!  & Train \\
			\hline 
			TensoRF & 15.68 & 0.455 & 0.658 & 25m29s & 15.46 & 0.609 & 0.566 & 29m07s \\
			Nerfacto & 19.05 & 0.549 & 0.495 & 11m37s & 18.29 & 0.688 & \cellcolor{yellow!40}0.422 & 11m16s \\ 
			+NeRFAcc & \cellcolor{yellow!40}20.14 & \cellcolor{orange!40}0.580 & \cellcolor{yellow!40}0.480 & 12m18s & \cellcolor{yellow!40}18.95 & \cellcolor{yellow!40}0.691 & 0.431 & 12m07s \\ 
			+Monocular Depth & 19.45 & 0.549 & 0.488 & 11m55s & 13.81 & 0.451 & 0.632 & 12m20s \\ 
			+Multi-view Depth & \cellcolor{orange!40}20.15 & \cellcolor{yellow!40}0.578 & \cellcolor{orange!40}0.465 & 12m24s & \cellcolor{orange!40}19.28 & \cellcolor{orange!40}0.706 & \cellcolor{orange!40}0.397 &  12m10s\\
			+Ours & \cellcolor{red!40}21.40 & \cellcolor{red!40}0.625 & \cellcolor{red!40}0.400 & 38m44s & \cellcolor{red!40}19.45 & \cellcolor{red!40}0.714 & \cellcolor{red!40}0.373 & 39m05s \\	
            \hline		
		\end{tabular}
	}
	\caption{
		\label{tab:comparisons} {Quantitative evaluation of our method compared to previous work, computed over two datasets. The '+' prefix indicates the included additional component to Nerfacto.}
	}
\end{table*}

\section{Experimental Results}

In this section, we demonstrate the effectiveness of our VS-NeRF approach, which includes a discussion of experimental settings and implementation details; evaluation results on benchmark datasets and comparison with previous work;
and an ablation study with respect to the major components in VS-NeRF.

\parag{Implementation Details.}

Our implementation of VS-NeRF is built upon the Nerfacto method from the Nerfstudio project~\citep{Tancik23}. It incorporates many published methods that have been found to work well for real data, such as Mip-NeRF360~\citep{Barron22}, IntantNGP~\citep{Muller22}, and NeRF-W~\citep{Martin21}.  We simply replace Nerfacto's sampling scheme by ours and add our depth-pushing loss and keep all other settings the same. Notably, the Nerfacto proposal network sampling scheme, from Mip-NeRF360~\citep{Barron22}, is also left unchanged. This ensures that any difference in performance is attributable to our regularization scheme. We turn off the camera optimization for both VS-NeRF and Nerfacto, since we observed a negative impact on datasets with accurate camera parameters. 

To reduce the time cost, we only activate the proposed view-consistent sampling technique in the first $5000$ iterations out of $30000$ in total, which is when the regularization of the geometry is the most needed. Empirically, the threshold $\delta$ in \cref{eq:vc_metric} is set to $0.4$, 
the weight of depth-pushing loss is set to $0.0001$, and the $\varepsilon$ in the depth pushing loss as in \cref{eq:depth_pushing_loss} is set to $0.01$.

\begin{table*}[ht]
    \centering
    \resizebox{0.85\linewidth}{!}{
		\begin{tabular}{@{}r@{\,\,}l@{\,\,}|ccc|@{\,}c@{\,}}
			\hline
			& & \!PSNR $\uparrow$\! & \!SSIM $\uparrow$\! & \!LPIPS $\downarrow$\! & Train  \\ \hline
			A) & Base Method Nerfacto & 18.88 & 0.544 & 0.503 & 13m20s \\
			B) & Base + VS (\cref{sec:sampling}) & 19.34 & 0.574 & 0.456 & 38m54s \\
			C) & Base + DL (\cref{sec:depth}) & 19.10 & 0.552 & 0.481 & 11m27s \\
			D) & Base + VS (Only Color Feature (\cref{eq:vc_metric})) + DL & 19.42 & 0.571 & 0.454 & 32m51s \\
			E) & Base + VS (Only Distilled Feature (\cref{eq:vc_metric})) + DL & 19.96 & 0.584 & 0.434 & 38m11s \\
			F) & Base + VS (Distilled Feature Dimension as $16$ (\cref{sec:distillation})) + DL  & \cellcolor{orange!40}20.04 & \cellcolor{orange!40}0.587 & \cellcolor{orange!40}0.432 & 39m22s \\
			G) & Base + VS (Distilled Feature Dimension as $64$ (\cref{sec:distillation})) + DL & \cellcolor{red!40}21.65 & \cellcolor{red!40}0.633 & \cellcolor{yellow!40}0.401 & 48m22s \\
			\hline 
			& Base + VS + DL (Our Complete Model) & \cellcolor{yellow!40}21.57 & \cellcolor{yellow!40}0.631 & \cellcolor{red!40}0.400 & 38m44s \\
			\hline
		\end{tabular}
	}
    \vspace{-3mm}
    \caption{
		\label{tab:ablations}
    	Ablation study in which we remove or replace the major components in our method to measure their effect on the Mip-NeRF360 dataset with $50$ training views. The two major components are VS: \textit{View-consistent Sampling} and DL: \textit{Depth-pushing Loss}.
    }
    \vspace{-2mm}
    \label{tab:ablation}
\end{table*}

\parag{Baselines.}

Since our implementation is based on \textbf{Nerfacto}~\citep{Tancik23}, we treat it as a baseline to demonstrate the positive impact of our adaptive sampling scheme and depth-pushing loss. We also use \textbf{TensoRF}~\citep{Chen22f} as another baseline that features efficient training. In addition we also compare against InstantNGP~\citep{Muller22}  from the Nerfstudio project. The most prominent difference between InstantNGP~\citep{Muller22} and \textbf{Nerfacto}~\citep{Tancik23} is the \textbf{NeRFAcc}~\citep{Li23d} efficient sampling scheme, whose name we will use to refer to this method.
We also incorporated our sampling method into MipNeRF360~\citep{Barron22}, which is significantly slower but yields better performance due to its  anti-aliasing design.

Regarding depth regularizations, we compare against both monocular and multi-view methods, using the depth-nerfacto method, again from Nerfstudio project. To test the method with \textbf{Monocular Depth} regularization, we use ZoeDepth~\citep{Bhat23} to predict pseudo depth and apply depth-ranking loss from SparseNeRF~\citep{Wang23a}. To test the method with \textbf{Multi-view Depth} regularization, we use the state-of-the-art MVSFormer++~\citep{Cao24} to provide depth estimations from correlating with adjacent $9$ views, along with the depth loss from DS-NeRF~\citep{Deng22c}. 

Additionally, we also apply our method to \textbf{Mip-NeRF360}~\citep{Barron22} and compare with \textbf{3DGS}~\citep{Kerbl23}.

\parag{Datasets and Metrics.} 

We use two benchmark datasets for our main evaluation, first the $9$ full scenes from Mip-NeRF360~\citep{Barron22} and second all $8$ scenes from the \textsc{Intermediate} official test set in Tanks $\&$ Temples dataset~\citep{Knapitsch2017}.
The scenes in the two datasets contain both a complex central object or area and a detailed background, and cover both bounded indoor scenes and large unbounded outdoor environments, making them challenging for NeRF methods. We use the same hyperparameter configuration for all experiments. 

In order to study the effect of the number of available views on the reconstruction quality, we subsample between $10$ to $110$ images per scene, $110$ being the size of the smallest image set in our datasets. To this end, for each scene, we first evenly sample $10$ images as an evaluation set and then sample evenly the remaining views. In ablation study, we use $50$ views for all scenes, as it is a reasonable number for practical usage and we observed that the need for regularization is highest as the number of views decreases.

Following the usual convention, we report quantitative results based on PSNR, SSIM~\citep{Wang04b}, and LPIPS~\citep{Zhang18h}, along with the training time in minutes as measured on a single NVIDIA A100 80GB GPU.

\subsection{Comparative Results} \label{main_exp}

\paragraph{Efficient NeRF Variants.} We report our comparative results with efficient NeRF variants on our two datasets as a function of the number of training views being used in \cref{fig:curve_comparisons}. We provide evaluation metrics averaged over all different scenes with different training view numbers in Tab.~\ref{tab:comparisons} and qualitative results in \cref{fig:comparisons}. VS-NeRF clearly outperforms all baselines in terms of novel view synthesis quality on Mip-NeRF360. On Tanks\&Temples, multi-view depth regularization is on par with our method when views are dense, but our method performs better in sparser cases. Crucially, our method significantly outperforms Nerfacto, upon which our implementation is based, which conclusively demonstrates the effectiveness of our view-consistent sampling and depth-pushing loss.  
Note that NeRFAcc also yields considerable improvement over Nefacto when available views are sparser but its relative performance drops when the number of views increases, which is not the case for our approach. TensoRF seems to struggle most, presumably due to the limitations of tensor-based scene representation in complex and unbounded scenes. We also compare agains two different depth-based regularizers, a monocular one using MVSformer++ and a multi-view one using ZoeDepth. Monocular depth estimation produces many artifacts and results in unstable performance, especially on Tanks\&Temples. This is largely because monocular depth is only a pseudo depth that may not be consistent across views. The multi-view depth regularization performs significantly better than the monocular one and consistently improves over Nerfacto. However, in challenging scenes with few available views it may produce unreliable depth estimations. In contrast, our method samples from a view-consistency distribution and is more robust as can be seen in results on Mip-NeRF360 dataset.

\begin{table}[h]
    \small
	\centering
	\scalebox{0.95}{
		\begin{tabular}{l|cccc}
			\hline
			Dataset & \multicolumn{4}{c}{Mip-NeRF360}  \\
			Method / Metric
			& \!PSNR$\uparrow$\!   & \!SSIM$\uparrow$\!    & \!LPIPS$\downarrow$\!  & Train \\
			\hline 
			Mip-NeRF360 & 22.54 & 0.642 & 0.347 & 29h22m \\
			Mip-NeRF360+Ours & 23.37 & 0.676 & 0.316 & 32h17m \\
			3DGS & 23.53 & 0.716 & 0.278 & 27m38s \\
            \hline		
		\end{tabular}
	}
	\caption{
		\label{tab:additional} {Additional results comparing to Mip-NeRF360 and 3D Gaussian Splatting.}
	}
	\vspace{-6mm}
\end{table}

\parag{Slower NeRF Variant.} In the above experiments, we compared against some of the most efficient NeRF variants. We also apply our method to slower Mip-NeRF360 and report our results in \cref{tab:additional}. Our sampling method clearly brings about an improvement. 

\vspace{-2mm}

\parag{Gaussian Splatting.}
For completeness, we also compare with 3D Gaussian Splatting~\citep{Kerbl23}, which has emerged as a powerful alternative to NeRF for novel view synthesis using an explicit point-based representation. To this end, we use the Splatfacto method from Nerfstudio~\citep{Tancik23}. 
As can be seen in \cref{tab:additional}, 3DGS yields the best performance of all for view synthesis. However, as discussed in \cref{sec:related}, there are many applications, from to surface reconstruction~\citep{Li23c} and 3D scene understanding~\citep{Kerr23} to inverse rendering~\citep{Jin23}, for which NeRF-like representations are more appropriate and yield better results than 3DGS. Thus, our results remain entirely relevant to these.

\vspace{-2mm}

\subsection{Ablation Study} \label{sec:ablations}

We perform an ablation study of the two novel components of our approach, i.e. the \textit{View-consistent Sampling} (VS) and the \textit{Depth-pushing Loss} (DL). For simplicity, the experiment is conducted with a moderate $50$-view setting for subsampling the scenes on Mip-NeRF360 dataset. The results are presented in Tab.~\ref{tab:ablations}. 
The first three rows of the table show that each component, VS or DL, brings an improvement when used separately. Comparing to our complete model in ie last row, it shows that they work best when used jointly.
Regarding the features used to compute the view-consistency metric, distilled DINOv2 features are more powerful than color feature when used independently, but the best performance is achieved by combining them, as in \cref{eq:vc_metric}. The dimensionality of the distilled features is also investigated here. We see that reducing the dimensionality to $16$ or increasing it to $64$ will degrade performance. Thus, we opt to use $32$ as the dimensionality of distilled features in our implementation.


\section{Conclusion}

We have proposed a novel view-consistent sampling technique as a regularization for the training of NeRF. The core idea is to combine high-level and low-level features to compute view-consistency metrics, and use it as a prior distribution
to sample on the ray. To mitigate the background collapse problem, we also propose a depth-pushing loss, which imposes a weaker regularization to favor distant samples on the ray. Extensive experiments on public datasets have demonstrated the effectiveness of the proposed method. Note that our method typically introduces more computational overhead than NeRF or Gaussian Splatting. In future work, we will look into more efficient implementations and extending our regularization method to Gaussian Splatting as well. 

{
    \small
    \bibliographystyle{ieeenat_fullname}
    \bibliography{main}

\begin{thebibliography}{65}
\providecommand{\natexlab}[1]{#1}
\providecommand{\url}[1]{\texttt{#1}}
\expandafter\ifx\csname urlstyle\endcsname\relax
  \providecommand{\doi}[1]{doi: #1}\else
  \providecommand{\doi}{doi: \begingroup \urlstyle{rm}\Url}\fi

\bibitem[Barron et~al.(2021)Barron, Mildenhall, Tancik, Hedman, Martin-Brualla,
  and Srinivasan]{Barron21}
J. Barron, B. Mildenhall, M. Tancik, P. Hedman, R. Martin-Brualla, and P.
  Srinivasan.
\newblock {Mip-Nerf: A Multiscale Representation for Anti-Aliasing Neural
  Radiance Fields}.
\newblock In \emph{International Conference on Computer Vision}, pages
  5855--5864, 2021.

\bibitem[Barron et~al.(2022)Barron, Mildenhall, Verbin, Srinivasan, and
  Hedman]{Barron22}
J. Barron, B. Mildenhall, D. Verbin, P.~P. Srinivasan, and P. Hedman.
\newblock {Mip-Nerf 360: Unbounded Anti-Aliased Neural Radiance Fields}.
\newblock In \emph{Conference on Computer Vision and Pattern Recognition},
  2022.

\bibitem[Barron et~al.(2023)Barron, Mildenhall, Verbin, Srinivasan, and
  Hedman]{Barron23}
Jonathan~T Barron, Ben Mildenhall, Dor Verbin, Pratul~P Srinivasan, and Peter
  Hedman.
\newblock Zip-nerf: Anti-aliased grid-based neural radiance fields.
\newblock In \emph{Proceedings of the IEEE/CVF International Conference on
  Computer Vision}, pages 19697--19705, 2023.

\bibitem[Bhat et~al.(2023)Bhat, Birkl, Wofk, Wonka, and M{\"u}ller]{Bhat23}
Shariq~Farooq Bhat, Reiner Birkl, Diana Wofk, Peter Wonka, and Matthias
  M{\"u}ller.
\newblock Zoedepth: Zero-shot transfer by combining relative and metric depth.
\newblock \emph{arXiv preprint arXiv:2302.12288}, 2023.

\bibitem[Cao et~al.(2024)Cao, Ren, and Fu]{Cao24}
Chenjie Cao, Xinlin Ren, and Yanwei Fu.
\newblock Mvsformer++: Revealing the devil in transformer's details for
  multi-view stereo.
\newblock \emph{arXiv preprint arXiv:2401.11673}, 2024.

\bibitem[Chen et~al.(2021)Chen, Xu, Zhao, Zhang, Xiang, Yu, and Su]{Chen21f}
Anpei Chen, Zexiang Xu, Fuqiang Zhao, Xiaoshuai Zhang, Fanbo Xiang, Jingyi Yu,
  and Hao Su.
\newblock Mvsnerf: Fast generalizable radiance field reconstruction from
  multi-view stereo.
\newblock In \emph{Proceedings of the IEEE/CVF International Conference on
  Computer Vision}, pages 14124--14133, 2021.

\bibitem[Chen et~al.(2022)Chen, Xu, Geiger, Yu, and Su]{Chen22f}
A. Chen, Z. Xu, A. Geiger, J. Yu, and H. Su.
\newblock {Tensorf: Tensorial Radiance Fields}.
\newblock In \emph{European Conference on Computer Vision}, pages 333--350,
  2022.

\bibitem[Deng et~al.(2022)Deng, Liu, Zhu, and Ramanan]{Deng22c}
Kangle Deng, Andrew Liu, Jun-Yan Zhu, and Deva Ramanan.
\newblock Depth-supervised nerf: Fewer views and faster training for free.
\newblock In \emph{Proceedings of the IEEE/CVF Conference on Computer Vision
  and Pattern Recognition}, pages 12882--12891, 2022.

\bibitem[Edstedt et~al.(2023)Edstedt, Sun, B{\"o}kman, Wadenb{\"a}ck, and
  Felsberg]{Edstedt23}
Johan Edstedt, Qiyu Sun, Georg B{\"o}kman, M{\aa}rten Wadenb{\"a}ck, and
  Michael Felsberg.
\newblock Roma: Revisiting robust losses for dense feature matching.
\newblock \emph{arXiv preprint arXiv:2305.15404}, 2023.

\bibitem[Fridovich-Keil et~al.(2022)Fridovich-Keil, Yu, Tancik, Chen, Recht,
  and Kanazawa]{Fridovich22}
S. Fridovich-Keil, A. Yu, M. Tancik, Q. Chen, B. Recht, and A. Kanazawa.
\newblock Plenoxels: Radiance fields without neural networks.
\newblock In \emph{Conference on Computer Vision and Pattern Recognition},
  pages 5501--5510, 2022.

\bibitem[Hartley and Zisserman(2000)]{Hartley00}
R. Hartley and A. Zisserman.
\newblock \emph{{Multiple View Geometry in Computer Vision}}.
\newblock Cambridge University Press, 2000.

\bibitem[He et~al.(2016)He, Zhang, Ren, and Sun]{He16a}
K. He, X. Zhang, S. Ren, and J. Sun.
\newblock {Deep Residual Learning for Image Recognition}.
\newblock In \emph{Conference on Computer Vision and Pattern Recognition},
  pages 770--778, 2016.

\bibitem[He et~al.(2022)He, Chen, Xie, Li, Doll{\'a}r, and Girshick]{He22a}
K. He, X. Chen, S. Xie, Y. Li, P. Doll{\'a}r, and R. Girshick.
\newblock {Masked Autoencoders Are Scalable Vision Learners}.
\newblock In \emph{Conference on Computer Vision and Pattern Recognition},
  pages 16000--16009, 2022.

\bibitem[Hertz et~al.(2021)Hertz, Perel, Giryes, Sorkine-Hornung, and
  Cohen-Or]{Hertz21}
Amir Hertz, Or Perel, Raja Giryes, Olga Sorkine-Hornung, and Daniel Cohen-Or.
\newblock Sape: Spatially-adaptive progressive encoding for neural
  optimization.
\newblock \emph{Advances in Neural Information Processing Systems},
  34:\penalty0 8820--8832, 2021.

\bibitem[Jain et~al.(2021)Jain, Tancik, and Abbeel]{Jain21}
Ajay Jain, Matthew Tancik, and Pieter Abbeel.
\newblock Putting nerf on a diet: Semantically consistent few-shot view
  synthesis.
\newblock In \emph{Proceedings of the IEEE/CVF International Conference on
  Computer Vision}, pages 5885--5894, 2021.

\bibitem[Jin et~al.(2023)Jin, Liu, Xu, Zhang, Han, Bi, Zhou, Xu, and Su]{Jin23}
Haian Jin, Isabella Liu, Peijia Xu, Xiaoshuai Zhang, Songfang Han, Sai Bi,
  Xiaowei Zhou, Zexiang Xu, and Hao Su.
\newblock Tensoir: Tensorial inverse rendering.
\newblock In \emph{Proceedings of the IEEE/CVF Conference on Computer Vision
  and Pattern Recognition}, pages 165--174, 2023.

\bibitem[Kerbl et~al.(2023)Kerbl, Kopanas, Leimk{\"u}hler, and
  Drettakis]{Kerbl23}
B. Kerbl, G. Kopanas, T. Leimk{\"u}hler, and G. Drettakis.
\newblock {3D Gaussian Splatting for Real-Time Radiance Field Rendering}.
\newblock \emph{ACM Transactions on Graphics}, 42\penalty0 (4), 2023.

\bibitem[Kerr et~al.(2023)Kerr, Kim, Goldberg, Kanazawa, and Tancik]{Kerr23}
J. Kerr, C. Kim, K. Goldberg, A. Kanazawa, and M. Tancik.
\newblock {LERF: Language Embedded Radiance Fields}.
\newblock In \emph{International Conference on Computer Vision}, 2023.

\bibitem[Kim et~al.(2022)Kim, Seo, and Han]{Kim22b}
Mijeong Kim, Seonguk Seo, and Bohyung Han.
\newblock Infonerf: Ray entropy minimization for few-shot neural volume
  rendering.
\newblock In \emph{Proceedings of the IEEE/CVF Conference on Computer Vision
  and Pattern Recognition}, pages 12912--12921, 2022.

\bibitem[Knapitsch et~al.(2017)Knapitsch, Park, Zhou, and
  Koltun]{Knapitsch2017}
Arno Knapitsch, Jaesik Park, Qian-Yi Zhou, and Vladlen Koltun.
\newblock Tanks and temples: Benchmarking large-scale scene reconstruction.
\newblock \emph{ACM Transactions on Graphics (ToG)}, 36\penalty0 (4):\penalty0
  1--13, 2017.

\bibitem[Kobayashi et~al.(2022)Kobayashi, Matsumoto, and Sitzmann]{kobayashi22}
S. Kobayashi, E. Matsumoto, and V. Sitzmann.
\newblock {Decomposing NeRF for Editing via Feature Field Distillation}.
\newblock In \emph{Advances in Neural Information Processing Systems}, 2022.

\bibitem[Li et~al.(2023{\natexlab{a}})Li, Gao, Tancik, and Kanazawa]{Li23d}
Ruilong Li, Hang Gao, Matthew Tancik, and Angjoo Kanazawa.
\newblock Nerfacc: Efficient sampling accelerates nerfs.
\newblock In \emph{Proceedings of the IEEE/CVF International Conference on
  Computer Vision}, pages 18537--18546, 2023{\natexlab{a}}.

\bibitem[Li and Snavely(2018)]{Li18o}
Zhengqi Li and Noah Snavely.
\newblock Megadepth: Learning single-view depth prediction from internet
  photos.
\newblock In \emph{Proceedings of the IEEE conference on computer vision and
  pattern recognition}, pages 2041--2050, 2018.

\bibitem[Li et~al.(2023{\natexlab{b}})Li, M\"uller, Evans, Taylor, Unberath,
  Liu, and Lin]{Li23c}
Z. Li, T. M\"uller, A. Evans, R. Taylor, M. Unberath, M. Liu, and C. Lin.
\newblock {Neuralangelo: High-Fidelity Neural Surface Reconstruction}.
\newblock In \emph{Conference on Computer Vision and Pattern Recognition},
  2023{\natexlab{b}}.

\bibitem[Luo et~al.(2024)Luo, Dunlap, Park, Holynski, and Darrell]{Luo24}
Grace Luo, Lisa Dunlap, Dong~Huk Park, Aleksander Holynski, and Trevor Darrell.
\newblock Diffusion hyperfeatures: Searching through time and space for
  semantic correspondence.
\newblock \emph{Advances in Neural Information Processing Systems}, 36, 2024.

\bibitem[Martin-Brualla et~al.(2021)Martin-Brualla, Radwan, Sajjadi, Barron,
  Dosovitskiy, and Duckworth]{Martin21}
Ricardo Martin-Brualla, Noha Radwan, Mehdi~SM Sajjadi, Jonathan~T Barron,
  Alexey Dosovitskiy, and Daniel Duckworth.
\newblock Nerf in the wild: Neural radiance fields for unconstrained photo
  collections.
\newblock In \emph{Proceedings of the IEEE/CVF Conference on Computer Vision
  and Pattern Recognition}, pages 7210--7219, 2021.

\bibitem[Mehta et~al.(2021)Mehta, Gharbi, Barnes, Shechtman, Ramamoorthi, and
  Chandraker]{Mehta21}
Ishit Mehta, Micha{\"e}l Gharbi, Connelly Barnes, Eli Shechtman, Ravi
  Ramamoorthi, and Manmohan Chandraker.
\newblock Modulated periodic activations for generalizable local functional
  representations.
\newblock In \emph{Proceedings of the IEEE/CVF International Conference on
  Computer Vision}, pages 14214--14223, 2021.

\bibitem[Mildenhall et~al.(2020)Mildenhall, P., Tancik, Barron, Ramamoorthi,
  and Ng]{Mildenhall20}
Ben Mildenhall, S.~P. P., M. Tancik, J.~T. Barron, R. Ramamoorthi, and R. Ng.
\newblock {NeRF: Representing Scenes as Neural Radiance Fields for View
  Synthesis}.
\newblock In \emph{European Conference on Computer Vision}, 2020.

\bibitem[M{\"u}ller et~al.(2022)M{\"u}ller, Evans, Schied, and
  Keller]{Muller22}
Thomas M{\"u}ller, Alex Evans, Christoph Schied, and Alexander Keller.
\newblock Instant neural graphics primitives with a multiresolution hash
  encoding.
\newblock \emph{ACM Transactions on Graphics (ToG)}, 41\penalty0 (4):\penalty0
  1--15, 2022.

\bibitem[Niemeyer et~al.(2022)Niemeyer, Barron, Mildenhall, Sajjadi, Geiger,
  and Radwan]{Niemeyer22}
Michael Niemeyer, Jonathan~T Barron, Ben Mildenhall, Mehdi~SM Sajjadi, Andreas
  Geiger, and Noha Radwan.
\newblock Regnerf: Regularizing neural radiance fields for view synthesis from
  sparse inputs.
\newblock In \emph{Proceedings of the IEEE/CVF Conference on Computer Vision
  and Pattern Recognition}, pages 5480--5490, 2022.

\bibitem[Oquab et~al.(2023)Oquab, Darcet, Moutakanni, Vo, Szafraniec, Khalidov,
  Fernandez, Haziza, Massa, El-Nouby, Howes, Huang, Xu, Sharma, Li, Galuba,
  Rabbat, Assran, Ballas, Synnaeve, Misra, Jegou, Mairal, Labatut, Joulin, and
  Bojanowski]{Oquab23}
M. Oquab, T. Darcet, T. Moutakanni, H. Vo, M. Szafraniec, V. Khalidov, P.
  Fernandez, D. Haziza, F. Massa, A. El-Nouby, R. Howes, P. Huang, H. Xu, V.
  Sharma, S. Li, W. Galuba, M. Rabbat, M. Assran, N. Ballas, G. Synnaeve, I.
  Misra, H. Jegou, J. Mairal, P. Labatut, A. Joulin, and P. Bojanowski.
\newblock {DINOv2: Learning Robust Visual Features Without Supervision}.
\newblock In \emph{arXiv Preprint}, 2023.

\bibitem[Prados and Faugeras(2005)]{Prados05}
E. Prados and O. Faugeras.
\newblock {Shape from Shading: A Well-Posed Problem?}
\newblock In \emph{Conference on Computer Vision and Pattern Recognition},
  2005.

\bibitem[Radford et~al.(2021)Radford, Kim, Hallacy, Ramesh, Goh, Agarwal,
  Sastry, Askell, Mishkin, Clark, et~al.]{Radford21}
A. Radford, J. Kim, C. Hallacy, A. Ramesh, G. Goh, S. Agarwal, G. Sastry, A.
  Askell, P. Mishkin, J. Clark, et~al.
\newblock Learning transferable visual models from natural language
  supervision.
\newblock In \emph{International Conference on Machine Learning}. PMLR, 2021.

\bibitem[Roessle et~al.(2022)Roessle, Barron, Mildenhall, Srinivasan, and
  Nie{\ss}ner]{Roessle22}
Barbara Roessle, Jonathan~T Barron, Ben Mildenhall, Pratul~P Srinivasan, and
  Matthias Nie{\ss}ner.
\newblock Dense depth priors for neural radiance fields from sparse input
  views.
\newblock In \emph{Proceedings of the IEEE/CVF Conference on Computer Vision
  and Pattern Recognition}, pages 12892--12901, 2022.

\bibitem[Sarlin et~al.(2020)Sarlin, DeTone, Malisiewicz, and
  Rabinovich]{Sarlin20}
P.E. Sarlin, D. DeTone, T. Malisiewicz, and A. Rabinovich.
\newblock {Superglue: Learning Feature Matching with Graph Neural Networks}.
\newblock In \emph{Conference on Computer Vision and Pattern Recognition},
  2020.

\bibitem[Shi et~al.(2024)Shi, Wei, Wang, and Su]{Shi24}
Ruoxi Shi, Xinyue Wei, Cheng Wang, and Hao Su.
\newblock Zerorf: Fast sparse view 360deg reconstruction with zero pretraining.
\newblock In \emph{Proceedings of the IEEE/CVF Conference on Computer Vision
  and Pattern Recognition}, pages 21114--21124, 2024.

\bibitem[Sitzmann et~al.(2020)Sitzmann, Martel, Bergman, Lindell, and
  Wetzstein]{Sitzmann20}
V. Sitzmann, J. Martel, A. Bergman, D. Lindell, and G. Wetzstein.
\newblock {Implicit Neural Representations with Periodic Activation Functions}.
\newblock In \emph{Advances in Neural Information Processing Systems}, 2020.

\bibitem[Somraj and Soundararajan(2023)]{Somraj23b}
Nagabhushan Somraj and Rajiv Soundararajan.
\newblock Vip-nerf: Visibility prior for sparse input neural radiance fields.
\newblock In \emph{ACM SIGGRAPH 2023 Conference Proceedings}, pages 1--11,
  2023.

\bibitem[Somraj et~al.(2023)Somraj, Karanayil, and Soundararajan]{Somraj23}
Nagabhushan Somraj, Adithyan Karanayil, and Rajiv Soundararajan.
\newblock Simplenerf: Regularizing sparse input neural radiance fields with
  simpler solutions.
\newblock In \emph{SIGGRAPH Asia 2023 Conference Papers}, pages 1--11, 2023.

\bibitem[Sun et~al.(2022)Sun, Sun, and Chen]{Sun22}
C. Sun, M. Sun, and H. Chen.
\newblock Direct voxel grid optimization: Super-fast convergence for radiance
  fields reconstruction.
\newblock In \emph{Conference on Computer Vision and Pattern Recognition},
  pages 5459--5469, 2022.

\bibitem[Sun et~al.(2021)Sun, Shen, Wang, Bao, and Zhou]{Sun21a}
J. Sun, Z. Shen, Y. Wang, H. Bao, and X. Zhou.
\newblock {LoFTR: Detector-Free Local Feature Matching with Transformers}.
\newblock In \emph{Conference on Computer Vision and Pattern Recognition},
  2021.

\bibitem[Tancik et~al.(2020)Tancik, Srinivasan, Mildenhall, Fridovich-Keil,
  Raghavan, Singhal, Ramamoorthi, Barron, and Ng]{Tancik20}
M. Tancik, P. Srinivasan, B. Mildenhall, S. Fridovich-Keil, N. Raghavan, U.
  Singhal, R. Ramamoorthi, J. Barron, and R. Ng.
\newblock {Fourier Features Let Networks Learn High Frequency Functions in Low
  Dimensional Domains}.
\newblock In \emph{Advances in Neural Information Processing Systems}, 2020.

\bibitem[Tancik et~al.(2023)Tancik, Weber, Ng, Li, Yi, Kerr, Wang,
  Kristoffersen, Austin, Salahi, Ahuja, McAllister, and Kanazawa]{Tancik23}
M. Tancik, E. Weber, E. Ng, R. Li, B. Yi, J. Kerr, T. Wang, A. Kristoffersen,
  J. Austin, K. Salahi, A. Ahuja, D. McAllister, and A. Kanazawa.
\newblock {Nerfstudio: A Modular Framework for Neural Radiance Field
  Development}.
\newblock In \emph{ACM SIGGRAPH}, 2023.

\bibitem[Truong et~al.(2023)Truong, Rakotosaona, Manhardt, and
  Tombari]{Truong23}
Prune Truong, Marie-Julie Rakotosaona, Fabian Manhardt, and Federico Tombari.
\newblock Sparf: Neural radiance fields from sparse and noisy poses.
\newblock In \emph{Proceedings of the IEEE/CVF Conference on Computer Vision
  and Pattern Recognition}, pages 4190--4200, 2023.

\bibitem[Turki et~al.(2024)Turki, Zollh{\"o}fer, Richardt, and
  Ramanan]{Turki24}
Haithem Turki, Michael Zollh{\"o}fer, Christian Richardt, and Deva Ramanan.
\newblock Pynerf: Pyramidal neural radiance fields.
\newblock \emph{Advances in Neural Information Processing Systems}, 36, 2024.

\bibitem[Tyszkiewic et~al.(2020)Tyszkiewic, Fua, and Trulls]{Tyszkiewic20}
M. Tyszkiewic, P. Fua, and E. Trulls.
\newblock {DISK: Learning Local Features with Policy Gradient}.
\newblock In \emph{Advances in Neural Information Processing Systems}, 2020.

\bibitem[Uy et~al.(2023)Uy, Martin-Brualla, Guibas, and Li]{Uy23}
Mikaela~Angelina Uy, Ricardo Martin-Brualla, Leonidas Guibas, and Ke Li.
\newblock Scade: Nerfs from space carving with ambiguity-aware depth estimates.
\newblock In \emph{Proceedings of the IEEE/CVF Conference on Computer Vision
  and Pattern Recognition}, pages 16518--16527, 2023.

\bibitem[Wang et~al.(2023{\natexlab{a}})Wang, Sun, Liu, Wu, Shen, Wu, Dai, and
  Zhang]{Wang23b}
Chen Wang, Jiadai Sun, Lina Liu, Chenming Wu, Zhelun Shen, Dayan Wu, Yuchao
  Dai, and Liangjun Zhang.
\newblock Digging into depth priors for outdoor neural radiance fields.
\newblock In \emph{Proceedings of the 31st ACM International Conference on
  Multimedia}, pages 1221--1230, 2023{\natexlab{a}}.

\bibitem[Wang et~al.(2023{\natexlab{b}})Wang, Chen, Loy, and Liu]{Wang23a}
Guangcong Wang, Zhaoxi Chen, Chen~Change Loy, and Ziwei Liu.
\newblock Sparsenerf: Distilling depth ranking for few-shot novel view
  synthesis.
\newblock In \emph{Proceedings of the IEEE/CVF International Conference on
  Computer Vision}, pages 9065--9076, 2023{\natexlab{b}}.

\bibitem[Wang et~al.(2021)Wang, Wang, Genova, Srinivasan, Zhou, Barron,
  Martin-Brualla, Snavely, and Funkhouser]{Wang21i}
Qianqian Wang, Zhicheng Wang, Kyle Genova, Pratul~P Srinivasan, Howard Zhou,
  Jonathan~T Barron, Ricardo Martin-Brualla, Noah Snavely, and Thomas
  Funkhouser.
\newblock Ibrnet: Learning multi-view image-based rendering.
\newblock In \emph{Proceedings of the IEEE/CVF Conference on Computer Vision
  and Pattern Recognition}, pages 4690--4699, 2021.

\bibitem[Wang et~al.(2004)Wang, Bovik, Sheikh, and Simoncelli]{Wang04b}
Zhou Wang, Alan~C Bovik, Hamid~R Sheikh, and Eero~P Simoncelli.
\newblock Image quality assessment: from error visibility to structural
  similarity.
\newblock \emph{IEEE transactions on image processing}, 13\penalty0
  (4):\penalty0 600--612, 2004.

\bibitem[Wei et~al.(2022)Wei, Fan, Xie, Wu, Yuille, and Feichtenhofer]{Wei22b}
Chen Wei, Haoqi Fan, Saining Xie, Chao-Yuan Wu, Alan Yuille, and Christoph
  Feichtenhofer.
\newblock Masked feature prediction for self-supervised visual pre-training.
\newblock In \emph{Proceedings of the IEEE/CVF Conference on Computer Vision
  and Pattern Recognition}, pages 14668--14678, 2022.

\bibitem[Wu et~al.(2023)Wu, Graikos, and Samaras]{Wu23}
Haoyu Wu, Alexandros Graikos, and Dimitris Samaras.
\newblock S-volsdf: Sparse multi-view stereo regularization of neural implicit
  surfaces.
\newblock In \emph{Proceedings of the IEEE/CVF International Conference on
  Computer Vision}, pages 3556--3568, 2023.

\bibitem[Wu et~al.(2024)Wu, Mildenhall, Henzler, Park, Gao, Watson, Srinivasan,
  Verbin, Barron, Poole, et~al.]{Wu2024}
Rundi Wu, Ben Mildenhall, Philipp Henzler, Keunhong Park, Ruiqi Gao, Daniel
  Watson, Pratul~P Srinivasan, Dor Verbin, Jonathan~T Barron, Ben Poole, et~al.
\newblock Reconfusion: 3d reconstruction with diffusion priors.
\newblock In \emph{Proceedings of the IEEE/CVF Conference on Computer Vision
  and Pattern Recognition}, pages 21551--21561, 2024.

\bibitem[Xie et~al.(2023)Xie, Geng, Hu, Zhang, Hu, and Cao]{Xie23}
Zhenda Xie, Zigang Geng, Jingcheng Hu, Zheng Zhang, Han Hu, and Yue Cao.
\newblock Revealing the dark secrets of masked image modeling.
\newblock In \emph{Proceedings of the IEEE/CVF conference on computer vision
  and pattern recognition}, pages 14475--14485, 2023.

\bibitem[Xu et~al.(2023)Xu, Guan, Wang, Liu, Zeng, Wang, and Yang]{Xu23e}
Luoyuan Xu, Tao Guan, Yuesong Wang, Wenkai Liu, Zhaojie Zeng, Junle Wang, and
  Wei Yang.
\newblock C2f2neus: Cascade cost frustum fusion for high fidelity and
  generalizable neural surface reconstruction.
\newblock In \emph{Proceedings of the IEEE/CVF International Conference on
  Computer Vision}, pages 18291--18301, 2023.

\bibitem[Yang et~al.(2023)Yang, Pavone, and Wang]{Yang23c}
Jiawei Yang, Marco Pavone, and Yue Wang.
\newblock Freenerf: Improving few-shot neural rendering with free frequency
  regularization.
\newblock In \emph{Proceedings of the IEEE/CVF conference on computer vision
  and pattern recognition}, pages 8254--8263, 2023.

\bibitem[Yi et~al.(2016)Yi, Trulls, Lepetit, and Fua]{Yi16b}
K.~M. Yi, E. Trulls, V. Lepetit, and P. Fua.
\newblock {LIFT: Learned Invariant Feature Transform}.
\newblock In \emph{European Conference on Computer Vision}, 2016.

\bibitem[Yu et~al.(2021)Yu, Ye, Tancik, and Kanazawa]{Yu21c}
Alex Yu, Vickie Ye, Matthew Tancik, and Angjoo Kanazawa.
\newblock pixelnerf: Neural radiance fields from one or few images.
\newblock In \emph{Proceedings of the IEEE/CVF Conference on Computer Vision
  and Pattern Recognition}, pages 4578--4587, 2021.

\bibitem[Yu et~al.(2022)Yu, Peng, Niemeyer, Sattler, and Geiger]{Yu22}
Zehao Yu, Songyou Peng, Michael Niemeyer, Torsten Sattler, and Andreas Geiger.
\newblock Monosdf: Exploring monocular geometric cues for neural implicit
  surface reconstruction.
\newblock \emph{Advances in neural information processing systems},
  35:\penalty0 25018--25032, 2022.

\bibitem[Zhang et~al.(2024)Zhang, Li, Yu, Huang, Gu, Zheng, and Bai]{Zhang24}
Jiawei Zhang, Jiahe Li, Xiaohan Yu, Lei Huang, Lin Gu, Jin Zheng, and Xiao Bai.
\newblock Cor-gs: sparse-view 3d gaussian splatting via co-regularization.
\newblock In \emph{European Conference on Computer Vision}, pages 335--352.
  Springer, 2024.

\bibitem[Zhang et~al.(2020)Zhang, Riegler, Snavely, and Koltun]{Zhang20f}
K. Zhang, G. Riegler, N. Snavely, and V. Koltun.
\newblock {Nerf++: Analyzing and Improving Neural Radiance Fields}.
\newblock In \emph{arXiv Preprint}, 2020.

\bibitem[Zhang et~al.(2018)Zhang, Isola, Efros, Shechtman, and Wang]{Zhang18h}
Richard Zhang, Phillip Isola, Alexei~A Efros, Eli Shechtman, and Oliver Wang.
\newblock {The Unreasonable Effectiveness of Deep Features as a Perceptual
  Metric}.
\newblock In \emph{Conference on Computer Vision and Pattern Recognition},
  pages 586--595, 2018.

\bibitem[Zhou et~al.(2021)Zhou, Wei, Wang, Shen, Xie, Yuille, and
  Kong]{Zhou21b}
Jinghao Zhou, Chen Wei, Huiyu Wang, Wei Shen, Cihang Xie, Alan Yuille, and Tao
  Kong.
\newblock ibot: Image bert pre-training with online tokenizer.
\newblock \emph{arXiv preprint arXiv:2111.07832}, 2021.

\bibitem[Zhu et~al.(2024)Zhu, He, Li, Li, and Chen]{Zhu24b}
Hanxin Zhu, Tianyu He, Xin Li, Bingchen Li, and Zhibo Chen.
\newblock Is vanilla mlp in neural radiance field enough for few-shot view
  synthesis?
\newblock In \emph{Proceedings of the IEEE/CVF Conference on Computer Vision
  and Pattern Recognition}, pages 20288--20298, 2024.

\end{thebibliography}
}

\end{document}